\documentclass[10pt,twocolumn,letterpaper]{article}

\usepackage{cvpr}              % To produce the CAMERA-READY version
\usepackage[dvipsnames, table]{xcolor}
\usepackage{minitoc}
\usepackage[accsupp]{axessibility}  % Improves PDF readability for those with disabilities.

\usepackage{gensymb}
\usepackage[normalem]{ulem}
\usepackage{tocloft}
\usepackage{balance}

\definecolor{cvprblue}{rgb}{0.21,0.49,0.74}
\usepackage[pagebackref,breaklinks,colorlinks,allcolors=cvprblue]{hyperref}

\def\paperID{****}
\def\confName{CVPR}
\def\confYear{2025}

\title{Wonderland: Navigating 3D Scenes from a Single Image}
\author{
  Hanwen Liang\textsuperscript{1,2*}\quad
  Junli Cao\textsuperscript{2,3*}\quad 
  Vidit Goel\textsuperscript{2}\quad
  Guocheng Qian\textsuperscript{2} \quad
  Sergei Korolev\textsuperscript{2} \\
  Demetri Terzopoulos\textsuperscript{3}\quad Konstantinos N. Plataniotis\textsuperscript{1}\quad
  Sergey Tulyakov\textsuperscript{2}\quad
  Jian Ren\textsuperscript{2} \\
  {\small {\textsuperscript{1}University of Toronto} \quad 
  {\textsuperscript{2}Snap Inc.} \quad
  {\textsuperscript{3}University of California, Los Angeles}}
}

\begin{document}

\twocolumn[{
\renewcommand\twocolumn[1][]{#1}

\maketitle
\begin{center}
    \captionsetup{type=figure}
    % \vspace{-6mm}
    \includegraphics[width=\textwidth]{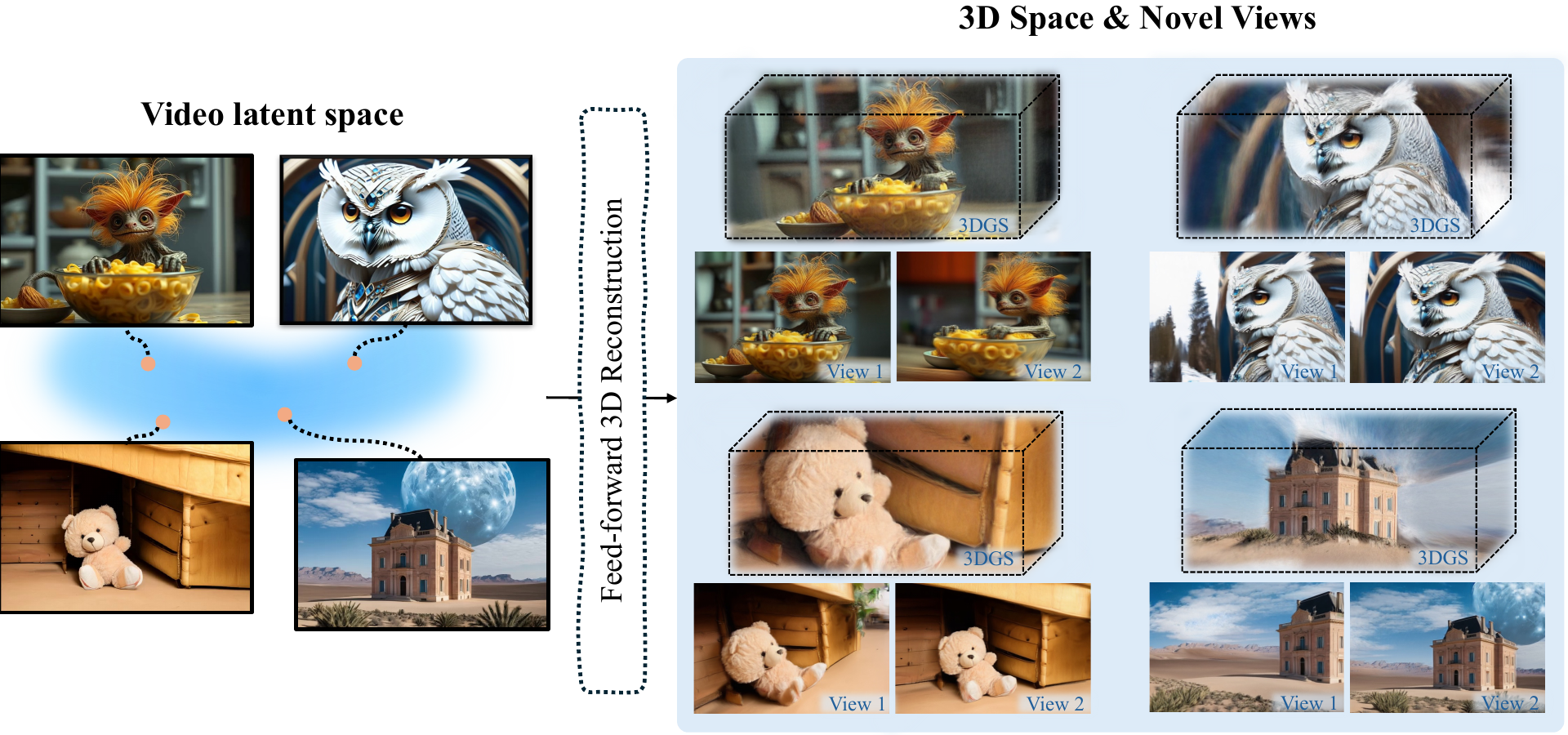}
    % \vspace{-6mm}
    \caption{\textbf{Visual results generated by Wonderland}. Given a single image, Wonderland reconstructs 3D scenes from the latent space of a camera-guided video diffusion model in a feed-forward manner.}
    \label{fig:teaser}
\end{center}
}]

\renewcommand{\thefootnote}{\fnsymbol{footnote}}
\footnotetext[1]{Equal contribution.}
\renewcommand{\thefootnote}{\arabic{footnote}}

\begin{abstract}
How can one efficiently generate high-quality, wide-scope 3D scenes from arbitrary single images?
Existing methods suffer several drawbacks, such as requiring multi-view data, time-consuming per-scene optimization, distorted geometry in occluded areas, and low visual quality in backgrounds.
Our novel 3D scene reconstruction pipeline overcomes these limitations to tackle the aforesaid challenge.
Specifically, we introduce a large-scale reconstruction model that leverages latents from a video diffusion model to predict 3D Gaussian Splattings of scenes in a feed-forward manner.
The video diffusion model is designed to create videos precisely following specified camera trajectories, allowing it to generate compressed video latents that encode multi-view information while maintaining 3D consistency.
We train the 3D reconstruction model to operate on the video latent space with a progressive learning strategy, enabling the efficient generation of high-quality, wide-scope, and generic 3D scenes.
Extensive evaluations across various datasets affirm that our model significantly outperforms existing single-view 3D scene generation methods, especially with out-of-domain images.
Thus, we demonstrate for the first time that a 3D reconstruction model can effectively be built upon the latent space of a diffusion model in order to realize efficient 3D scene generation.
Project page:\\ {\small\url{https://snap-research.github.io/wonderland/}}
\end{abstract}    

\vspace{-0.3cm}
\section{Introduction}

Humans have an innate ability to perceive 3D information from \emph{single} images, effortlessly estimating distances, interpreting spatial arrangements, and mentally filling in occluded or neighboring regions of wide-scope 3D scenes.
Replicating this capability with learnable algorithms, however, is extremely difficult because of the limited geometric and contextual information in a single view.
Recent advancements in learnable scene representations (\eg, Neural Radiance Fields (NeRF)~\cite{nerf} and 3D Gaussian Splatting (3DGS)~\cite{3dgs}) show promise for rendering photo-realistic 3D scenes, but they suffer two major limitations that hinder their scalability and flexibility: Their training requires dense \emph{multi-view images} and they employ a time-consuming \emph{per-scene optimization} scheme.

To mitigate the need for multi-view data, several studies~\citep{viewcrafter, cat3d, reconx, zero123, zeronvs} use generative priors in image diffusion models~\cite{ho2020denoising,song2020score,latentdiffusion} for 3D synthesis from a single view or sparse views.
However, these methods suffer from limited 3D consistency in novel view synthesis; \eg, incorrect or distorted geometry in occluded regions~\cite{lucid,realdreamer,zeronvs} and blurry backgrounds~\cite{cat3d, reconfusion}, indicating that image diffusion models have limited abilities in 3D contexts.
They lack the spatio-temporal modeling mechanisms and knowledge priors necessary to maintain geometric coherence across multiple views~\cite{liang2022self}.
Recent work~\cite{reconx, viewcrafter} utilizes video diffusion models conditioned on point clouds constructed from sparse views, but they are confined to narrow scopes and struggle with generating regions invisible from the input views, due to irregular black artifacts in conditional frames rendered from incomplete local point clouds.

To avoid lengthy optimization processes for building 3D scene representations, recent efforts explore generalizable models to perform feedforward reconstruction~\citep{lrm, lgm, xu2024grm, pixelsplats, jin2024lvsm, ziwen2024long, feng2024geogs3d, li2023instant3d, szymanowicz2024splatter}.
State-of-the-art approaches~\citep{pixelsplats, xu2024grm, gslrm, liang2024feed} use transformers to regress 3D Gaussian parameters from sparse posed images;
however, they face significant memory challenges, as training and the rendering of complex scenes involve handling vast numbers of tokens encoded from overlapping high-resolution input views.
Consequently, existing methods are limited mostly to object-level generation or scenes with narrow view angles and scope~\cite{lrm, gslrm, xu2024controlrm, szymanowicz2024flash3d} requiring fewer input views and less computation.

\begin{figure*}
  \centering
  \includegraphics[width=1.0\linewidth]{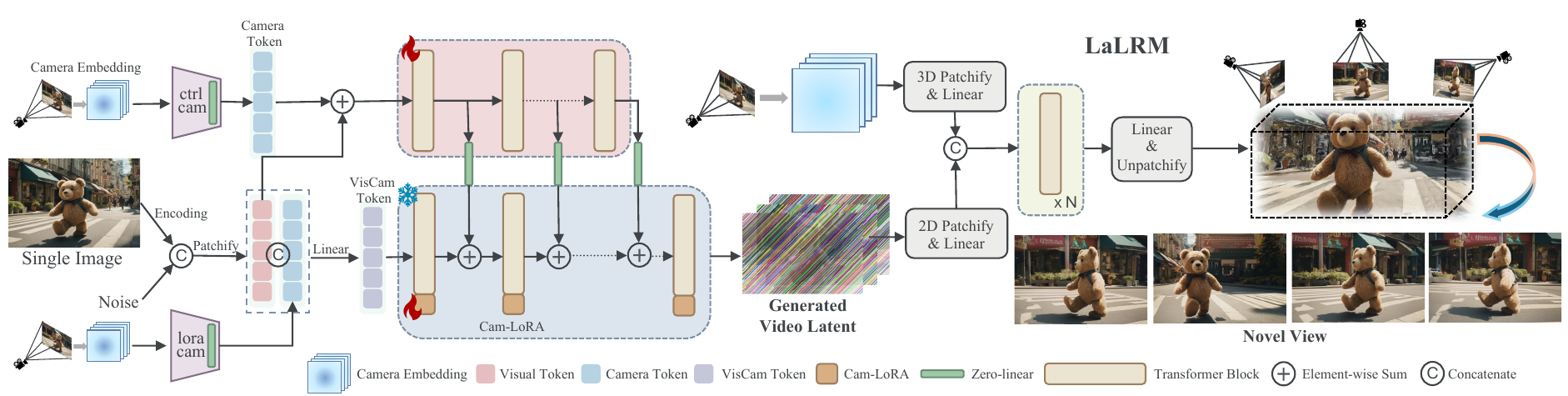}
  \caption{\textbf{Overview of Wonderland.} Given a single image, a camera-guided video diffusion model follows the camera trajectory and generates a 3D-aware video latent. This is leveraged by the Latent Large Reconstruction Model (LaLRM) to construct the 3D scene in a feed-forward manner. The video diffusion model incorporates dual-branch camera conditioning to achieve precise pose control. The LaLRM operates in the video latent space and efficiently reconstructs a wide-scope, high-fidelity 3D scene.}
  \label{fig:method}
  % \vspace{-5mm}
\end{figure*}

This paper introduces \textit{Wonderland}, which effectively addresses the above two challenges.
Referring to \autoref{fig:method}, from a single image Wonderland can \emph{efficiently} generate \emph{high-fidelity}, \emph{wide-scope} 3D scenes (see, \eg, \autoref{fig:teaser}).
We explore rich 3D scene understanding instilled in video diffusion models and build a point-based 3D representation (\ie, a 3DGS~\cite{3dgs}) regressed from video latents in a feed-forward manner, dramatically reducing memory requirements and accelerating the reconstruction process.
To realize these benefits, we propose the following techniques:
\begin{enumerate}
    \item We introduce a representation for \emph{controllable} 3D generation by leveraging generative priors from \emph{camera-guided video diffusion models}.
    Unlike image diffusion models, video diffusion models are trained on extensive video datasets, capturing comprehensive spatial relationships within scenes from multiple views and imbuing the latent space with a form of ``3D awareness'', which enables 3D consistency in novel view synthesis.
    \item To achieve controllable novel view generation, we empower the video models with precise control over specified camera motions, by introducing a novel dual-branch conditioning mechanism that effectively incorporates the desired \emph{diverse} camera trajectories into the video diffusion model, thus enabling it to expand a single image into a multi-view consistent capturing of the 3D scene.
    \item For efficient 3D reconstruction, we propose a novel Latent Large Reconstruction Model (LaLRM) that regresses the 3DGS from video latents.
    Our model lifts generated video latents to 3D space in a \emph{feed-forward manner}, effectively aligning the generation and reconstruction tasks—and bridging image space and 3D space—through the video latent space.
    Compared with reconstructing scenes from images (\ie, decoded videos) the video latents offer a $256\times$ spatiotemporal reduction, while retaining essential and consistent 3D structural details.
    Such high compression is crucial, as it allows the LaLRM to handle a wider range of high-fidelity 3D scenes within the reconstruction framework, subject to the same memory constraints.  
\end{enumerate}
We extensively evaluate Wonderland and verify that it achieves state-of-the-art performance in 3D scene generation with the following unique advantages:
\begin{itemize}
\item Leveraging the dual-branch camera conditioning strategy, our video diffusion model generates 3D-consistent multi-view capturing of scenes, affording more precise pose control than prior methods (\autoref{tab:camera_table} and \autoref{fig:camera_comparison}).

\item In the zero-shot novel view synthesis setting, by using a single image as input for feed-forward 3D scene reconstruction, our approach outperforms prior methods on various benchmark datasets; \ie, RealEstate10K~\cite{zhou2018stereo}, DL3DV~\cite{ling2024dl3dv}, and Tanks-and-Temples~\cite{Knapitsch2017} (\autoref{tab:3d_table}). 

\item By operating on the latent space, our reconstruction pipeline harnesses the generative capabilities of the video diffusion model, enabling it to render \emph{high-quality} images, \emph{wide-scope} views, and more \emph{generic} and \emph{diverse} (\eg, out-of-domain) scenes that go far beyond object-level reconstruction (\autoref{fig:recon} and \autoref{fig:inthewild_fig}).

\end{itemize}

\section{Related Work}
\label{sec:relatedwork}

\textbf{3D Scene Generation:}
Object-level 3D generation~\cite{liu2024one, Magic123, shi2023mvdream, DreamFusion, voleti2024sv3d, chan2023generative, chen2023fantasia3d, feng2024geogs3d,tochilkin2024triposr} has made significant progress, whereas 3D scene generation lags behind.
Most approaches follow a two-stage pipeline: generating novel views from sparse inputs, then per-scene optimizing a 3D representation. 
Early methods~\cite{lucid, realdreamer, yu2024wonderjourney} combine depth-based warping with diffusion-based inpainting, but monocular depth estimation and per-view inpainting often introduce distortions and artifacts, compromising 3D consistency.
Recent efforts~\cite{hollein2024viewdiff, cat3d, reconfusion} incorporate camera conditioning into image diffusion models to control poses in novel view generation, yet they still suffer from blurry backgrounds.
More recent approaches utilize video diffusion models and global point clouds to improve multi-view consistency~\cite{viewcrafter,reconx}, but their effectiveness depends on point cloud quality and remains limited to narrow-scoped scenes.
Also, these previous methods use time-consuming per-scene optimization.
By contrast, our approach integrates explicit camera control into a video diffusion model for expansive 3D scene generation with precise pose control and we develop a large reconstruction model to efficiently construct 3D scenes directly from video latents in a feed-forward manner.

\textbf{Camera-Conditioned Video Diffusion Models} have attracted much attention~\cite{cheong2024boosting, xiao20243dtrajmaster, li2025realcam, bahmani2024ac3d, wang2024cpa, xiao2024trajectory, he2025cameractrl, yuan2024generative, jin2025flovd, kuang2024collaborative,schwarz2025generative}.
Earlier efforts explore training-free strategies~\cite{hu2024motionmaster, hou2024training} or apply LoRA~\cite{AnimateDiff, blattmann2023stable, hu2021lora, sun2024dimensionx} modules in video diffusion models for specific-type camera control. 
Later research explores injecting the camera matrices into the video diffusion models~\cite{wang2024motionctrl,xu2024camco, he2024cameractrl}, yet these approaches can degrade the quality of visual generation.
Recently, VD3D~\cite{bahmani2024vd3d} adopts ControlNet-like~\cite{zhang2023adding} conditioning with a cross-attention mechanism, but because of the high computational cost, pose control is integrated only to a low-resolution video generator in a cascaded generation pipeline.
By comparison, we devise a dual-branch conditioning applied directly to the video diffusion foundation, which achieves precise pose control without degrading the generation quality.

\textbf{Sparse-View Reconstruction:}
Using vanilla NeRF or 3DGS for reconstruction has strict requirements for dense multi-view consistent images. Some approaches mitigate the dense capture requirement~\cite{nerf, DietNeRF, sparsenerf, TrackNeRF}, yet they are slow to optimize. Recent efforts in feed-forward 3D reconstruction have attracted more interest~\citep{pixelnerf, chen2024mvsplat, gslrm, lrm}. Training on large-scale 3D datasets, these methods use transformers to regress 3D representations directly from sparse input views. Although effective, the computational overhead and number of views (\eg, $2-6$ views) limit them to narrower tasks; \eg, object-level reconstructions. 
We alleviate these limitations by operating in the latent space of video diffusion models, which enables our approach to work with a large number of views while keeping computational efficiency in check. 
\section{Method}

In this section, we present our framework for 3D scene generation conditioned on a single image.
First, we develop a camera-guided video diffusion transformer to generate video latents covering a wide scope of the scene.
Precise pose control is achieved with our novel dual-branch camera conditioning module.
The generated video latents are compact and 3D-aware, since they encapsulate a multi-view capturing of the scene consistent in both structure and appearance, making them ideal to be lifted up to 3D space. 
Second, we develop a novel Latent Large Reconstruction Model (LaLRM) to lift the video latents to 3D Gaussian Splatting (3DGS)~\cite{3dgs} for scene construction in a feed-forward and memory-efficient manner.

\subsection{Preliminaries}

\textbf{Latent Video Diffusion Transformers:}
\label{sec:pre_video}
Recent video diffusion models work in latent space~\cite{cogvideo,liu2024sora,blattmann2023stable,chen2023videocrafter1} for efficiency and use transformer-based architectures for scalability and quality~\cite{brooks2024video, peebles2023scalable, menapace2024snap}.
Specifically, given a source video $x\in R^{T\times H\times W\times 3}$ with $T$ RGB frames of spatial dimension $H\times W$, an encoder $\mathcal{E}$ from 3D-VAE~\cite{liu2024sora, cogvideo} first compresses it to a latent $z \in R^{t\times h\times w\times c}$, where $t$, $h$, $w$, and $c$ are temporal length, height, width, and channel dimension. The spatial and temporal compression rates are defined as $r_s=\frac{H}{h}=\frac{W}{w}$ and $r_t=\frac{T}{t}$.
During training, a noisy latent is created in the forward diffusion process as $z_{\tau}=\alpha_{\tau}z + \sigma_{\tau}\epsilon$, where $\epsilon \sim \mathcal{N}(0,I)$, and $\alpha$ and $\sigma$ depend on the noise scheduler parameterized via a diffusion-time $\tau$. 
Then, $z_\tau$ is passed to the transformer model $D_{\theta}$, parameterized by $\theta$, that first patchifies $z_\tau$ into visual tokens as a long sequence $o_v\in R^{N_v\times d_v}$ of length $N_v$ and dimensionality $d_v$.
Later, $o_v$ is passed through a series of transformer blocks.
The transformer output is unpatchified and projected to restore the original latent shape for the loss computation, formulated as
\begin{equation}
\mathcal{L}_{\theta} \triangleq E_{x\sim p_\text{data}, \tau\sim p_{\tau}}\left[{\left\| D_{\theta}(z_{\tau}; y,\tau) -\epsilon\right\|}_{2}^{2}\right],\\[-2mm]
\end{equation}
where $y$ is the conditional signal.\footnote{For a list of mathematical notation, refer to \autoref{Tab:notation} in the supplemental document comprising Appendices A through E.}
The exact objective may vary depending on the model's parameterization~\cite{blattmann2023stable, cogvideo}.

\textbf{3D Gaussian Splatting:} We use 3DGS as the scene representation for its visual fidelity and rendering speed~\cite{3dgs}.
It represents 3D scene appearance and geometry as a collection of 3D Gaussians (parameterized by mean position, $3\times3$ covariance matrix, color, and opacity).

\subsection{Camera-Guided Video Latent Generation}
\label{sec:camera-video-model}

Video diffusion models have made significant progress in generating high-quality, physically plausible visualizations of scenes.
However, using them to synthesize 3D-aware latents is nontrivial, since they lack explicit control over pose trajectory and may produce dynamic scenes that are unsuitable for downstream 3D reconstruction. 
To address this challenge, we enhance video generation models with \textit{precise camera control} and adapt them to \textit{generate static scenes}, which enables them to perform controlled and comprehensive explorations of static scenes and produce 3D-aware latents suitable for 3D scene reconstruction.

\textbf{Camera Representation:}
To achieve precise pose control, instead of using frame-level camera parameters, we enrich condition information with more fine-grained pixel-level positional representations; \ie, the Plücker embedding~\cite{griffiths2014principles}.
Given the camera parameters of frame $f$ consisting of rotation $\mathbf{R}_f$, translation $\mathbf{t}_f$, and intrinsic $\mathbf{K}_f$, the Plücker coordinate of the pixel at position ($u_f, v_f$) in frame $f$ is $\dot{p}_{u_f, v_f}=(\mathbf{t}_f\times d'_{u_f, v_f}, d'_{u_f, v_f}) \in R^6$, where $\times$ denotes the cross-product and $d'_{(\cdot)}={d_{(\cdot)}}/{\left\|d_{(\cdot)} \right\|}$ is the normalized ray direction $d_{u_f, v_f}=\mathbf{R}_{f}\mathbf{K}_{f}^{-1}[u_f, v_f, 1]^{T} + \mathbf{t}_f$.
For each video $x$, computing pixel-level Plücker coordinates yields the Plücker embedding $p \in R^{T\times H\times W\times 6}$, which comprises the spatiotemporal camera pose information.

\textbf{Dual Branch Camera Guidance:}
It is challenging to adapt a pretrained video diffusion transformer to generate static scenes that precisely follow the camera trajectory while preserving high visual quality.
The difficulty arises from the entangled attention mechanisms in spatiotemporal visual tokens, where even minor architectural modifications can degrade visual quality~\cite{bahmani2024vd3d}.
Moreover, fully fine-tuning on small-scale static scene datasets is computationally intensive and risks overfitting thus compromising the generalized knowledge acquired from web-scale pre-training.
To address these challenges, we devise a dual-branch conditioning mechanism inspired by the principles of ControlNet~\cite{zhang2023adding} and LoRA~\cite{hu2021lora}, which are renown for their ability to extend pretrained models without altering their original weights.
ControlNet is effective in integrating conditions and LoRA is efficient in finetuning diffusion models on customized datasets with reduced overfitting risk.

As illustrated in \autoref{fig:method}, the Plücker embedding $p$ of the camera is fed into two lightweight camera encoders to obtain two sets of camera tokens, $o_\text{ctrl}$ and $o_\text{lora}$.
Both camera encoders incorporate 3D Convolution (3DConv) layers to spatiotemporally compress $p$ and an unfolding operation, resulting in camera tokens of the same dimension as the video tokens $o_v$.
Motivated by prior work~\cite{chen2023pixart,zhang2023adding}, the camera encoders are followed by a zero-linear layer ($\mathcal{F}_{d_v, d_v}$) for stable training at the initial stage.

To build the ControlNet branch, we create a trainable copy of the first $N$ base transformer blocks of the base video model, trading off between controllability and memory costs.
The $o_\text{ctrl}$ is element-wise added to $o_v$ and fed into the first trainable block.
The output of trainable block $i$ is connected to a zero-linear layer and then element-wise added to the output of the corresponding frozen block, facilitating an effective and in-depth integration of camera conditions.
To further enhance camera control, camera information is injected into the main branch with LoRA finetuning. 
We concatenate $o_\text{lora}$ and $o_v$ channel-wise and feed them into a tailored linear layer ($\mathcal{F}_{2\times d_v, d_v}$).
The weights $W_\mathcal{F}\in R^{2d_v, d_v}$ of this layer are initialized with an identity matrix in the $o_v$ component, making the output identical to $o_v$ at the initial training stage.
The output is sent into frozen transformer blocks in the main branch, where the trainable camera-LoRA module is learned~\cite{cogvideo}, enabling model fine-tuning with minimal computational cost.
The LoRA module also adapts the model to the customized static-scene training sets, enhancing the overall static nature in the generated scenes.
The effects of the ControlNet and LoRA branches are analyzed in \autoref{tab:camera_table} and \autoref{sec:abl_video}.

\subsection{Latent Large Reconstruction Model}

From the video latents, our goal at this stage is to enable fast, high-quality GS reconstruction of 3D scenes with wide-view coverage.
We adopt a transformer-based architecture to regress pixel-aligned Gaussian primitives in a feed-forward manner.
A straightforward design is to predict GS from pixels.
However, in our settings involving wide-scope coverage with high-dimensional decoded videos, prior image encoding mechanisms (\eg, $8\times8$ patchification in~\cite{gslrm}) leads to over 260K tokens, which is computationally unaffordable for dense transformers with quadratic complexity.
Increasing the patch size can alleviate this burden, but the aggressive compression over the decoded video risks losing fine-grained details~\cite{caron2021emerging, he2022masked} and impairing reconstruction performance.

Motivated by the success of latent-based diffusion models~\cite{blattmann2023align, latentdiffusion}, we propose the Latent Large Reconstruction Model (LaLRM), largely reducing the time cost of the per-scene optimization strategy.
The LaLRM operates on the latent space with three key advantages.
First, the video latent provides a highly compressed representation, which is essential for enabling wide-scope 3D scene reconstruction under memory constraints.
Second, as the autoencoder is pretrained with perceptual~\cite{zhang2018unreasonable} and patch-based~\cite{isola2017image} adversarial objectives~\cite{yu2021vector, dosovitskiy2016generating}, the video latent retains a perceptually equivalent representation as images~\cite{latentdiffusion}, which facilitates faithful, detailed, and high-quality reconstructions.
Third, pretraining on web-scale datasets makes the video latent a robust representation, allowing the LaLRM to generalize effectively to out-of-domain scenes.
Thus, developed upon the latent space, our model can efficiently reconstruct \textit{high-quality, wide-scope, and generic} 3D scenes.

\textbf{Latent-based 3D Reconstruction Model:}
Given the video latent $z\in R^{t\times h\times w\times c}$ and the corresponding camera poses in the Plücker embedding $p \in R^{T\times H\times W\times 6}$, we first transform them into latent tokens and pose tokens.
We design patchify modules to ensure that both token sets have identical lengths.
They are channel-wise concatenated, linearly projected to a lower channel dimension, and then fed into a chain of base transformer blocks.
We devise a lightweight decoding module that uses the output tokens to regress the attributes of Gaussians.
We fulfill pixel-level correspondence between the Gaussians and source video $x$ in RGB space via the decoding module, which involves a 3D DeConv layer with a 12-channel output (3 RGB, 3 scale, 4 rotation quaternion, 1 opacity, and 1 ray distance) yielding a Gaussian feature map $G\in R^{(T\times H\times W)\times 12}$ whose features correspond to 3D Gaussian features.
\autoref{supp:implementation-details} provides additional architectural design details.
During training, we render images from the predicted Gaussians by randomly choosing $V$ supervision views and minimizing the total image reconstruction loss $\mathcal{L}_\text{recon} = \lambda_{1}\mathcal{L}_\text{mse} + \lambda_{2}\mathcal{L}_\text{perc}$, a weighted combination of the mean squared error loss $\mathcal{L}_\text{mse}$ and perceptual loss $\mathcal{L}_\text{perc}$ based on a VGG-19 network~\cite{simonyan2014very}. 

\textbf{Progressive Training Strategy:}
Due to the large domain gap between the video latent and the 3DGS representation, training our reconstruction model faces challenges:
\begin{itemize}
    \item To achieve 3D geometry construction and avoid overfitting to the seen views (those that are directly decoded from the video latent by the 3DVAE), we must involve more unseen views to guarantee 3D consistency.
    \item To facilitate generic 3D scene generation by the LaLRM, the reconstruction model should adapt to in-the-wild video latents generated by the video diffusion model.
    \item Aiming for superior visual quality we must reconstruct the scene in high resolution.
\end{itemize}
Therefore, we adopt a progressive training strategy in regard to the data source and image resolution.
We initiate the model training at a lower video resolution using benchmark video datasets with known camera poses. 
Each video has abundant frames sufficient for seen view and unseen view sampling. 
We use a large stride $s$ to sample a video clip, covering a scene range of $s\times T$ sampled frames, which are considered \textit{seen views}.
The remaining frames within the scene range and the frames outside the scene range are considered \textit{unseen views}.
The VAE encoder $\mathcal{E}$ projects the video clip to the video latent, which is forwarded into our reconstruction model for GS prediction.
We use $V'$ seen views and $V-V'$ unseen views for supervision during training.
Then, we scale the training up to higher resolutions.
At this stage, we incorporate both benchmark datasets and a substantial volume of \emph{out-of-domain} videos generated by our own camera-guided video diffusion model.
For these generated samples, we use the video latents along with their conditioned camera poses as inputs, while the decoded video frames from these latents provide supervision views.
The progressive training enables our model to better generalize to out-of-domain videos, enhancing robustness and fidelity in 3D reconstruction.
\begin{figure*}
% \vspace{-5mm}
  \centering
  \includegraphics[width=1.0\linewidth]{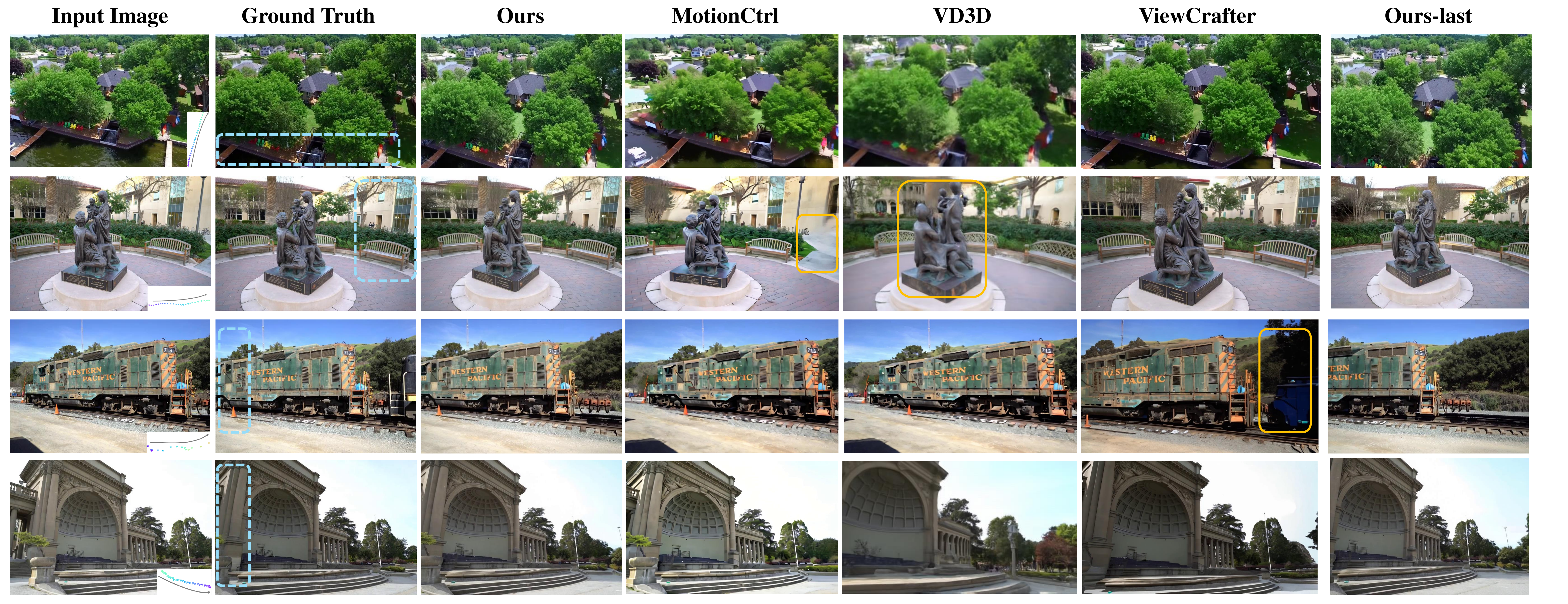}
  % \vspace{-5mm}
  \caption{\textbf{Qualitative comparison against prior arts in camera-guided video generation}. 
  Frame 14 in each sample is shown for comparison, with the first column displaying the conditional image and camera trajectory (bottom-right).
  Blue bounding boxes denote reference areas to assist comparison and orange bounding boxes highlight low-quality generations.
  We also show our last frames in the rightmost column. 
  Our method outperforms the priors in both precise camera control and high-quality and wide-scope video generation.}
  \label{fig:camera_comparison}
  % \vspace{-4mm}
\end{figure*}

\section{Experiments}

\textbf{Implementation Details:}
As the base model to build our camera-guided video generation model, we employed a transformer-based video diffusion model, CogVideoX-5B-I2V~\citep{cogvideo} that generates 49 frames of $480\times720$ pixels resolution.
A 3DVAE~\citep{cogvideo} was used to compress the video clips with temporal and spatial ratios of $r_t=4$ and $r_s=8$, producing latents of dimension $13\times 60\times 90$.
For the ControlNet branch, we used the first $N=21$ base transformer blocks from the video model to initialize the weights.
The camera-LoRA has a low rank of dimension $256$.
To build the transformer architecture for the LaLRM, we used 24 base transformer blocks with a hidden dimension of 1,024. 
The model was first trained on low-resolution video clips ($49\times 240\times 360$) and then finetuned at higher resolution ($49\times 480\times 720$).
A total of $V=48$ supervision views were used, for which we randomly selected $V'=24$ frames from the source video clip as seen views and an additional 24 frames disjoint from the sampled video frames as unseen views. 
See \autoref{supp:implementation-details} for additional details.

\textbf{Training Datasets:}
We utilized three benchmark datasets with camera pose annotations to train our models, RealEstate10K (RE10K)~\citep{zhou2018stereo}, ACID~\citep{liu2021infinite}, and DL3DV~\citep{ling2024dl3dv}.
RE10K consists of around 80K videos, primarily capturing static real estate environments. % at a frame rate of 30fps.
We used a dynamic stride $s\in \{3,4,5\}$ to sample video clips, covering wide scene ranges of approximately 150 to 250 frames.
ACID contains videos of natural landscapes, with 11K scenes in the training set and 20K in the test set, and 
as most contain fewer than 100 frames with annotated camera poses, we set $s\in \{1,2\}$.
DL3DV, comprising training split DL3DV-10K and test split DL3DV-140, includes diverse real-world indoor and outdoor scenes, with
200 to 300 keyframes per video plus camera pose annotations.
Due to the dramatic view changes between consecutive keyframes, we sampled clips using stride $s=1$.
We utilized only the standard training splits during model deployment.
For the out-of-domain data used to train the LaLRM, we generated 20K videos with image prompts from Flux.1~\citep{flux_repo} and camera poses sampled from RE10K. Due to the large scene scale differences across the datasets, we normalized the camera poses following reference~\citep{zhou2018stereo}.

\begin{table}
\centering
    \caption{\textbf{Quantitative comparison against prior art in camera-guided video generation} on RealEstate10K, DL3DV, and Tanks and Temples dataset. We report the performance for visual quality (FID and FVD), camera-guidance precision ($R_{\text{err}}$ and $T_{\text{err}}$), and visual similarity (LPIPS, PSNR, and SSIM).}
    \label{tab:camera_table}
    \resizebox{\columnwidth}{!}{
    \begin{tabular}{lccccccc}
    \toprule        
        Method & \multicolumn{7}{c}{Metrics} \\
         \cmidrule(lr){2-8}
        \textit{Dataset} & FID $\downarrow$ & FVD $\downarrow$ & $R_{\text{err}}$ $\downarrow$ & $T_{\text{err}}$ $\downarrow$ & LPIPS $\downarrow$ & PSNR $\uparrow$ & SSIM $\uparrow$ \\
        \midrule
        \textit{RealEstate10K} & \multicolumn{7}{c}{}  \\
        MotionCtrl~\cite{wang2024motionctrl} & 22.58 & 229.34  & 0.231 & 0.794 & 0.296 & 14.68 & 0.402 \\
        VD3D~\cite{bahmani2024vd3d} & 21.40 & 187.55 & 0.053 & 0.126 & 0.227 & 17.26 & 0.514 \\
        ViewCrafter~\cite{viewcrafter} & 20.89 & 203.71 & 0.054 & 0.152 & 0.212 & 18.91 & 0.501 \\
        \rowcolor[rgb]{ .816,  .808,  .808} Ours & \textbf{16.16} & \textbf{153.48} & \textbf{0.046} & \textbf{0.093} & \textbf{0.206} & \textbf{19.71} & \textbf{0.557} \\
        \midrule
        \textit{DL3DV} & \multicolumn{7}{c}{}  \\
        MotionCtrl~\cite{wang2024motionctrl} & 25.58 & 248.77 & 0.467 & 1.114 & 0.309 & 14.35 & 0.385 \\
        VD3D~\cite{bahmani2024vd3d} & 22.70 & 232.97 & 0.094 & 0.237 & 0.259 & 16.28 & 0.487 \\
        ViewCrafter~\cite{viewcrafter} & 20.55 & 210.62 & 0.092 & 0.243 & 0.237 & 17.10 & 0.519 \\
        \rowcolor[rgb]{ .816,  .808,  .808} Ours & \textbf{17.74} & \textbf{169.34} & \textbf{0.061} & \textbf{0.130} & \textbf{0.218} & \textbf{17.56} & \textbf{0.543} \\
        \midrule
        \textit{Tanks and Temples} & \multicolumn{7}{c}{}  \\
        MotionCtrl~\cite{wang2024motionctrl} & 30.17 & 289.62 & 0.834 & 1.501 & 0.312 & 14.58 & 0.386  \\
        VD3D~\cite{bahmani2024vd3d} & 24.33 & 244.18 & 0.117 & 0.292 & 0.284 & 15.35 & 0.467\\
        ViewCrafter~\cite{viewcrafter} & 22.41 & 230.56 & 0.125 & 0.306 & 0.245 & 16.20 & 0.506 \\
        \rowcolor[rgb]{ .816,  .808,  .808} Ours & \textbf{19.46} & \textbf{189.32} & \textbf{0.094} & \textbf{0.172} & \textbf{0.221} & \textbf{16.87} & \textbf{0.529}\\ 
        \midrule     
        Ablations on \textit{RE10K} & \multicolumn{7}{c}{}  \\
        Lora-branch & 19.02 & 212.74 & 0.102 & 0.157 & - &- &- \\
        Ctrl-branch & 18.75 & 205.45 & 0.058 & 0.104 &  - &- &- \\
        Dual-branch & \textbf{17.22} & \textbf{183.54} & \textbf{0.052} & \textbf{0.095} &  - &- & - \\       
        \bottomrule
    \end{tabular}}
    \vspace{-5mm}
\end{table}

\subsection{Camera-Guided Video Generation}
\label{sec:cam_compare}

We evaluated our camera-guided video generation model (\autoref{sec:camera-video-model}) by comparing the visual generation quality and camera-guidance precision against three baselines; \ie, MotionCtrl~\citep{wang2024motionctrl}, VD3D~\citep{bahmani2024vd3d}, and ViewCrafter~\citep{viewcrafter}.

\textbf{Evaluation Datasets}:
\begin{itemize}
    \item RE10K~\citep{zhou2018stereo} --- we randomly selected 300 videos from the test set. For each video, a starting frame and the subsequent $n$ camera poses are sampled as image and pose conditions, $n$ being determined by the video length.
    \item DL3DV-140~\citep{ling2024dl3dv} --- we randomly sampled 300 video clips from this dataset using the same strategy as above.
    \item Tanks-and-Temples (Tanks)~\citep{Knapitsch2017} --- we randomly sampled 100 video clips from all 14 scenes using the same strategy as above. These were employed to test out-of-domain generalization. We used COLMAP~\citep{schoenberger2016mvs,schoenberger2016sfm} to annotate poses for the source videos as they lack dense pose annotations for all frames.
\end{itemize}

\begin{figure*}
%  \vspace{-5mm}
  \centering
  \includegraphics[width=0.9\linewidth]{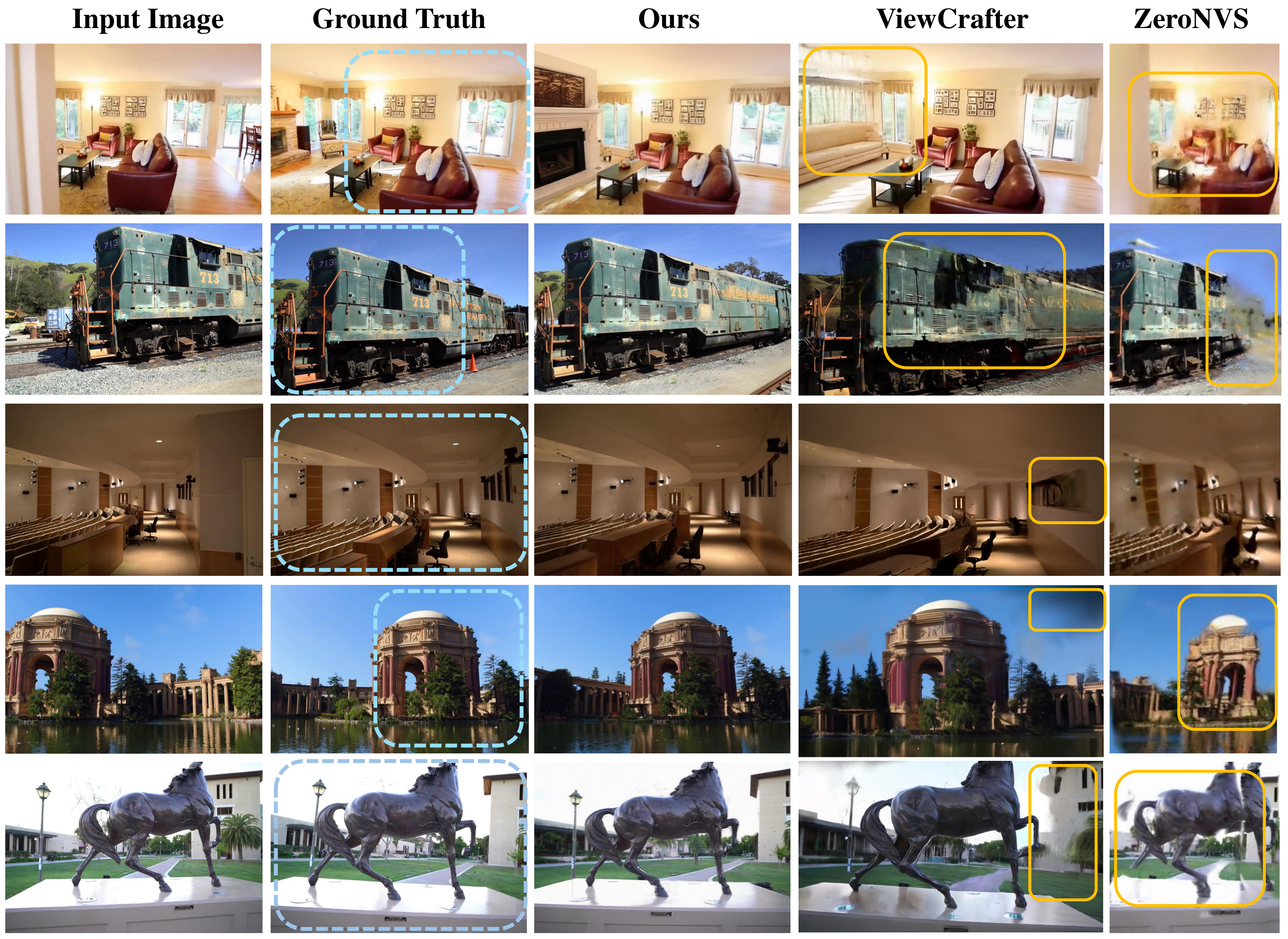}
  % \vspace{-5mm}
  \caption{\textbf{Qualitative comparison of 3D scene generation}. Blue bounding boxes show visible regions from conditional images and yellow bounding boxes show low-quality regions. Our approach generates much higher quality novel views from \emph{one} conditional image. Note that ZeroNVS generations have a square resolution.}
  \label{fig:recon}
 % \vspace{-5mm}
\end{figure*}

\textbf{Evaluation Metrics:} We compared the generated videos with ground-truth video clips using multiple metrics.
\begin{itemize}
    \item The \textit{Visual Quality and Temporal Coherence} of the generated videos were evaluated using Fréchet Inception Distance (FID)~\citep{heusel2017gans} and Fréchet Video Distance (FVD)~\citep{fvd}. 
    
    \item The \textit{Camera-Guidance Precision} was measured using Rotation error ($R_\text{err}$) and Translation error ($T_\text{err}$) metrics. Following~\citep{wang2024motionctrl,he2024cameractrl,xu2024camco,bahmani2024vd3d}, camera poses of generated videos were obtained via COLMAP~\citep{schoenberger2016sfm}, followed by converting the camera systems to be relative to the first frame and normalizing all cameras into a unified scale~\citep{bahmani2024vd3d}.
    Since the different baseline videos vary in length, for a fair comparison, we measured the mean errors across the frames generated under the same pose conditions.
    
    \item The \textit{Visual Similarity} was assessed by calculating PSNR, SSIM~\citep{ssim}, and LPIPS~\citep{zhang2018unreasonable} between the generated images and ground-truth views. For a reliable comparison, we evaluated all methods over the first 14 frames as the generated videos tend to deviate from the conditional single view and present diverse appearances as the scene progresses, making similarity metrics, which measure differences from ground-truth views, less reliable for evaluating generation quality.

\end{itemize}

\begin{figure*}
% \vspace{-5mm}
  \centering
  \includegraphics[width=1.0\linewidth]{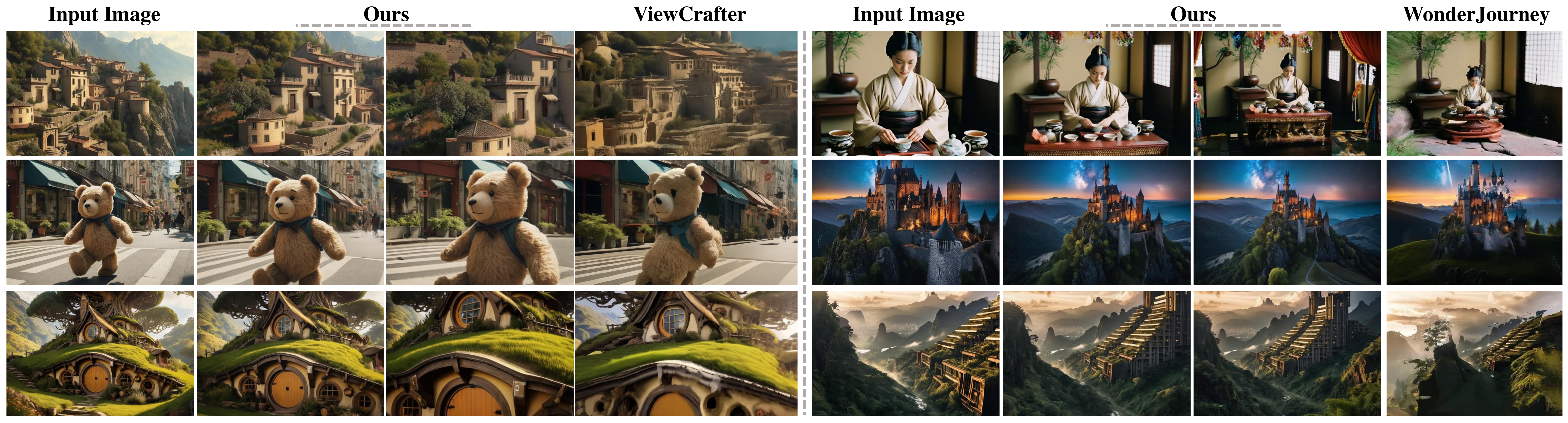}
  % \vspace{-6mm}
  \caption{Comparison with ViewCrafter (\emph{left}) and WonderJourney (\emph{right}) for in-the-wild 3D scene generation from \emph{single} input images. }
  \label{fig:inthewild_fig}
  \vspace{-2mm}
\end{figure*}

\textbf{Qualitative Comparisons} are presented in \autoref{fig:camera_comparison}.
Although MotionCtrl produces images with good quality, it fails in precise alignment with the given camera conditions because it adopts a high-level camera embedding to control camera poses, which lacks fine-grained pose control.
The frames generated by VD3D have limited quality due to the low resolution.
ViewCrafter suffers artifacts across frames, mainly because it uses incomplete point clouds to render images as conditions, which have irregular missing regions.
By contrast, our method demonstrates superior quality for precise pose control and higher visual fidelity.

\textbf{Quantitative Comparisons}
are shown in \autoref{tab:camera_table}. Our method consistently outperforms the baselines across all metrics.
The smaller FID and FVD indicate that our model follows the ground-truth data distribution more precisely.
The smaller LPIPS and larger PSNR and SSIM indicate better visual quality and greater similarity to the ground truth.
Our method is also capable of generating more precise camera control than the baselines (\ie, lower $R_\text{err}$ and $T_\text{err}$).

\subsection{3D Scene Generation}

We evaluated our method and two baseline approaches (\ie, ZeroNVS~\citep{zeronvs} and ViewCrafter~\citep{viewcrafter}) in 3D scene generation on real-world and in-the-wild synthetic datasets.
ViewCrafter generates 3D scenes conditioned on a single image and a pose trajectory, whereas ZeroNVS reconstructs 360-degree scenes without any pose conditioning.

\textbf{Comparison of Benchmark Datasets:} To evaluate 3D scene generation on the benchmark datasets, we sampled 100, 100, and 50 images along with camera trajectories from the RE10K, DL3DV, and Tanks test sets, respectively, using the sampling strategy described in \autoref{sec:cam_compare}.
For a quantitative comparison, we measured the LPIPS, SSIM, and PSNR by assessing the renderings against ground-truth frames from the source video.
Evaluating in this under-constrained setting is challenging, as multiple 3D scenes can be regarded as consistent generations for a given view~\citep{cat3d}.
Therefore, following \autoref{sec:cam_compare}, we measured the metrics using frames that are \emph{temporally close} to the conditional image; \ie, a total of 14 sampled frames and poses subsequent to the conditional image.

\begin{table}
% \vspace{-5mm}
\centering
    \caption{\textbf{Quantitative comparison} on various benchmark datasets for 3D scene novel view synthesis with \emph{single view} condition.}
    \label{tab:3d_table}
    \resizebox{\columnwidth}{!}{%
    \begin{tabular}{lccccccccc}
    \toprule   
    Method & \multicolumn{3}{c}{RealEstate10K} & \multicolumn{3}{c}{DL3DV} & \multicolumn{3}{c}{Tanks-and-Temples} \\
         \cmidrule(lr){2-4} \cmidrule(lr){5-7} \cmidrule(lr){8-10}
        \textit{Metrics} & LPIPS $\downarrow$ & PSNR $\uparrow$ & SSIM $\uparrow$ & LPIPS $\downarrow$ & PSNR $\uparrow$ & SSIM $\uparrow$ & LPIPS $\downarrow$ & PSNR $\uparrow$ & SSIM $\uparrow$ \\
        \midrule
        ZeroNVS~\cite{zeronvs} & 0.448 & 13.01 & 0.378 & 0.465 & 13.35 & 0.339 & 0.470 & 12.94 & 0.325 \\
        ViewCrafter~\cite{viewcrafter} & 0.341 & 16.84 & 0.514 & 0.352 & 15.53 & 0.525 & 0.384 & 14.93 & 0.483 \\
        \rowcolor[rgb]{ .816,  .808,  .808} Ours & \textbf{0.292} & \textbf{17.15} & \textbf{0.550} & \textbf{0.325} & \textbf{16.64} & \textbf{0.574} & \textbf{0.344} & \textbf{15.90} & \textbf{0.510} \\
        \bottomrule
    \end{tabular}
    }
\vspace{-0.5cm}
\end{table}

The qualitative comparison shown in \autoref{fig:recon} reveals the superior 3D generation capabilities of our model. ZeroNVS produces renderings that are noticeably blurry and lack details. ViewCrafter shows improved results in regions visible from the conditional image while failing to handle occluded areas properly. By contrast, our model excels at preserving intricate details and accurately reconstructing the visible regions in the conditional image. By leveraging priors from the video diffusion model, our approach also generates high-fidelity and visually satisfactory novel views, even in unseen regions. \autoref{tab:3d_table} provides a quantitative comparison, which reveals that our method outperforms all baselines by a large margin across multiple datasets. 
These results affirm that our model can generate high-fidelity, geometrically consistent 3D scenes from single views.

\textbf{Comparison of In-the-Wild Scene Generation:}
We generated synthetic images using SDXL~\citep{podell2023sdxl}, and randomly sampled trajectories from RE10K dataset for novel view synthesis.
Here, we focused on wide scene generation, performing qualitative comparisons against ViewCrafter~\citep{viewcrafter} and Wonderjourney~\citep{yu2024wonderjourney} because of their prowess in expansive 3D scene synthesis.
Since Wonderjourney generates scenes exclusively by zooming away from the conditional image, our comparisons with Wonderjourney were confined to this scenario. 
With ViewCrafter, we performed comparisons on more complex pose trajectories to evaluate varied and wide-scope scene generation.

\autoref{fig:inthewild_fig} shows that ViewCrafter can generate 3D scenes only within a very limited area, and the quality deteriorates significantly when the view range grows large. While Wonderjourney can produce scenes with a broader field of view, the generated views tend to be blurry and they contain many artifacts. Comparatively, our method generates expansive scenes that maintain high realism and are consistent in both appearance and 3D geometry.

\begin{figure}
  \centering
  \includegraphics[width=1.0\linewidth]{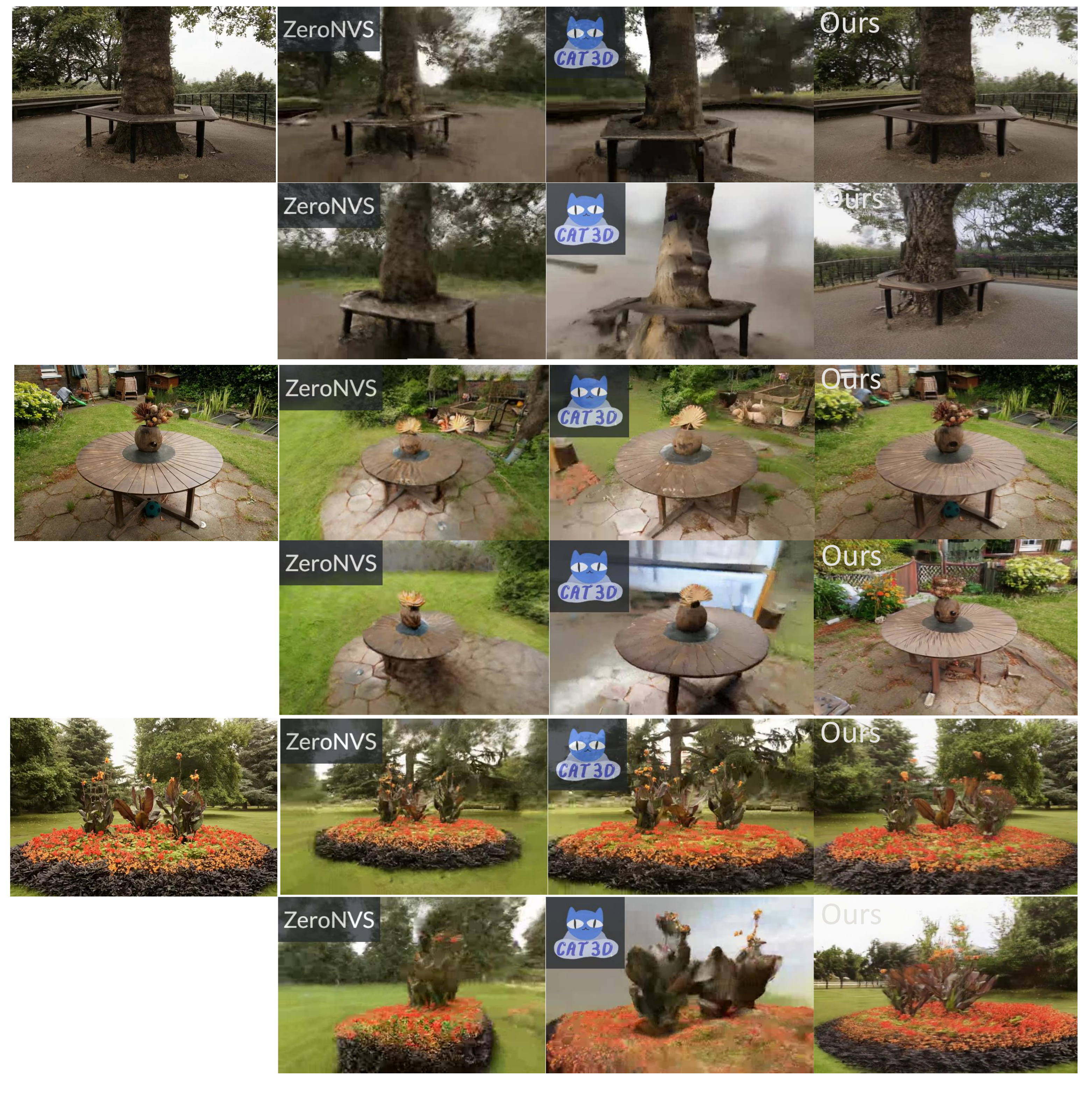}
  \caption{\textbf{Comparison of ZeroNVS and Cat3D on the Mip-Nerf dataset} in 3D scene generation from \emph{single} input images. For each scene, the conditional image is shown in the left column along with renderings from two viewpoints, one at the conditional image (starting) view (upper) and another at around a 120\degree rotation from the starting view (lower).}
  \vspace{-0.3cm}
  \label{fig:cat3d}
\end{figure}

\paragraph{Comparison of Mip-NeRF:}
We next compare our method with Cat3D~\citep{cat3d} on more complex scenes from Mip-NeRF~\citep{barronmip}. Due to the lack of open-source code for Cat3D, we retrieved the demonstration results directly from the source webpage of reference \citep{cat3d}.
The images in \autoref{fig:cat3d} were rendered from 3DGS representations generated with orbital camera trajectories. For each scene, we show renderings from two viewpoints: one at the conditional image (starting) view and another rotated at around 120\degree from the starting view.
We observe that for views close to the conditional image, our method achieves rendering quality similar to Cat3D and noticeably better than ZeroNVS.
However, as the viewpoint deviates from the conditional image, Cat3D suffers from severe blurring, particularly in the background.
By contrast, our method generates scenes with clearer textures, sharper details, and greater consistency with the conditional images.

\textbf{Comparison of Latency:} To highlight the efficiency of our processing pipeline, we report end-to-end latency on a single NVIDIA A100 GPU and compare it against recent work: Cat3D~\citep{cat3d} requires approximately 16 minutes ($1$\,min with 16 A100 GPUs), ViewCrafter~\citep{viewcrafter} takes over 6 minutes (25-frame video), and ZeroNVS~\citep{zeronvs} consumes an extensive 3 hours. By contrast, our method completes scene generation in around 5 minutes, achieving a $3.2\times$ speedup over Cat3D, $1.2\times$ over ViewCrafter, and $36\times$ over ZeroNVS, thus affirming our pipeline's superiority in single-image 3D scene generation.

\subsection{Analysis of the Architectural Design}
\label{sec:abl_dualbranch}

\textbf{Dual-Branch Camera-Pose Guidance:}
To analyze the effect of each conditioning branch in the camera-pose guidance video diffusion model, we developed two single-branch models: 
\emph{Ctrl-branch}, which has the ControlNet-based conditioning, and \emph{Lora-branch}, which has feature concatenation and LoRA-based finetuning.
We sampled 100 video clips from the RE10K test set and measured the metrics FID, FVD, $R_\text{err}$, and $T_\text{err}$. The final part of \autoref{tab:camera_table} reports the comparison, showing that compared to \emph{Lora-branch}, \emph{Ctrl-branch} contributes more for precise pose control. Adding \emph{Lora-branch} on top of \emph{Ctrl-branch}, which yields \emph{Dual-branch}, further improves the controllability and generation quality of the scene. 
Additional ablation analyses are found in \autoref{sec:abl_video}.

\textbf{Latents \emph{vs.} RGB for 3D Reconstruction:}
We compared the performance of using video latents from 3DVAE versus RGB frames for 3D reconstruction.
All the designs take source videos as input but vary in their encoding mechanisms.
Two RGB-based reconstruction models use normal patchifications for encoding and adopt the same transformer backbone as our LaLRM: 
(a) \textit{RGB-49} samples 49-frame clips using the same sampling strategy as the LaLRM.
(b) \textit{RGB-14} samples 14-frame clips with a sampling stride increased by four to cover a similar scene range as the LaLRM.
To ensure comparable and affordable computational and memory cost, we introduced a 3DConv layer and a 2DConv layer at the top of the RGB-49 and RGB-14 models, respectively. They patchify the RGB inputs into a lower-dimensional feature space, matching the input latents in the LaLRM.
Additionally, we developed another latent-based reconstruction model (c), in which the 3DVAE~\citep{cogvideo} encoder is used to embed images and is finetuned along with the reconstruction transformer.
The training strategies remain consistent across these designs.
We evaluated the models on RE10K (100 test samples) for the in-domain and on Tanks (50 test samples) for the out-of-domain tests.
The sampled video clips were fed into the models via various encoding mechanisms, and the transformer constructed the 3DGS scenes. The unsampled frames within the view coverage are considered unseen views for evaluation purposes.
\autoref{tab:lalrm_ablation} shows that our LaLRM encodes frames via frozen-encoder and performs best relative to the others.
Reconstruction model (c) exhibits degraded performance, suggesting that joint training compromises the robust latent representation from the original 3DVAE.

\begin{table}
    \centering
    \caption{Ablations on RGB-based (rgb-14/rgb-49) and Latent-based (encoder-finetuned/encoder-frozen) reconstruction designs. LaLRM (with encoder frozen) produces superior results.}
    \label{tab:lalrm_ablation}
    \resizebox{\columnwidth}{!}{
    \begin{tabular}{lcccccc}
    \toprule
 Method & \multicolumn{3}{c}{RealEstate10K} & \multicolumn{3}{c}{Tanks-and-Temples} \\
         \cmidrule(lr){2-4} \cmidrule(lr){5-7}
        \textit{Metrics} & LPIPS $\downarrow$ & PSNR $\uparrow$ & SSIM $\uparrow$ & LPIPS $\downarrow$ & PSNR $\uparrow$ & SSIM $\uparrow$ \\
        \midrule
        RGB-based & 0.137/0.126 & 21.39/25.06 & 0.751/0.830 & 0.221/0.192 & 19.70/20.54 & 0.605/0.687 \\
        Latent-based & 0.129/\textbf{0.122} & 26.14/\textbf{27.10} & 0.841/\textbf{0.864} & 0.201/\textbf{0.170} & 20.75/\textbf{22.66} & 0.716/\textbf{0.743} \\
        \bottomrule
    \end{tabular}
    }
    \vspace{-0.2cm}
\end{table}

\section{Conclusions}

We have introduced \textit{Wonderland}, a novel framework for high-fidelity 3D scene generation from \emph{single} images.
It addresses the multi-view requirement and time-consuming optimization issue for scalable 3D scene synthesis, by leveraging the rich generative priors embedded in camera-conditioned video diffusion models and operating within the compact latent space for feed-forward reconstruction.
A \emph{dual-branch} camera conditioning mechanism enables precise pose control and static nature in novel view synthesis. 
The latent large reconstruction model integrates the generative capabilities of video diffusion models with 3D Gaussian Splatting, ensuring computational efficiency and scalability to wide-view coverage. 
Our extensive evaluations across diverse datasets have demonstrated the superior performance of our approach in generating high-quality, wide-scope, and generic 3D scenes.
It outperforms existing state-of-the-art methods in both video generalization and 3D rendering quality.

\section{Acknowledgment}
We would like to thank Hsin-Ying Lee, Chaoyang Wang, and Peiye Zhuang for valuable discussions, insights and contributions in the development of this work.
% \clearpage
\balance
\bibliographystyle{IEEEtranS}
{\small\bibliography{main}}

% Generated by IEEEtranS.bst, version: 1.14 (2015/08/26)
\begin{thebibliography}{100}
\providecommand{\url}[1]{#1}
\csname url@samestyle\endcsname
\providecommand{\newblock}{\relax}
\providecommand{\bibinfo}[2]{#2}
\providecommand{\BIBentrySTDinterwordspacing}{\spaceskip=0pt\relax}
\providecommand{\BIBentryALTinterwordstretchfactor}{4}
\providecommand{\BIBentryALTinterwordspacing}{\spaceskip=\fontdimen2\font plus
\BIBentryALTinterwordstretchfactor\fontdimen3\font minus \fontdimen4\font\relax}
\providecommand{\BIBforeignlanguage}[2]{{%
\expandafter\ifx\csname l@#1\endcsname\relax
\typeout{** WARNING: IEEEtranS.bst: No hyphenation pattern has been}%
\typeout{** loaded for the language `#1'. Using the pattern for}%
\typeout{** the default language instead.}%
\else
\language=\csname l@#1\endcsname
\fi
#2}}
\providecommand{\BIBdecl}{\relax}
\BIBdecl

\bibitem{bahmani2023cc3d}
S.~Bahmani, J.~J. Park, D.~Paschalidou, X.~Yan, G.~Wetzstein, L.~Guibas, and A.~Tagliasacchi, ``{CC3D}: Layout-conditioned generation of compositional {3D} scenes,'' in \emph{Proceedings of the IEEE/CVF International Conference on Computer Vision}, 2023, pp. 7171--7181.

\bibitem{bahmani2024ac3d}
S.~Bahmani, I.~Skorokhodov, G.~Qian, A.~Siarohin, W.~Menapace, A.~Tagliasacchi, D.~B. Lindell, and S.~Tulyakov, ``{AC3D}: Analyzing and improving {3D} camera control in video diffusion transformers,'' \emph{arXiv preprint arXiv:2411.18673}, 2024.

\bibitem{bahmani20244d}
S.~Bahmani, I.~Skorokhodov, V.~Rong, G.~Wetzstein, L.~Guibas, P.~Wonka, S.~Tulyakov, J.~J. Park, A.~Tagliasacchi, and D.~B. Lindell, ``{4D-fy}: Text-to-{4D} generation using hybrid score distillation sampling,'' in \emph{Proceedings of the IEEE/CVF Conference on Computer Vision and Pattern Recognition}, 2024, pp. 7996--8006.

\bibitem{bahmani2024vd3d}
S.~Bahmani, I.~Skorokhodov, A.~Siarohin, W.~Menapace, G.~Qian, M.~Vasilkovsky, H.-Y. Lee, C.~Wang, J.~Zou, A.~Tagliasacchi \emph{et~al.}, ``{VD3D}: Taming large video diffusion transformers for {3D} camera control,'' \emph{arXiv preprint arXiv:2407.12781}, 2024.

\bibitem{barronmip}
J.~T. Barron, B.~Mildenhall, D.~Verbin, P.~P. Srinivasan, and P.~Hedman, ``{Mip-NeRF} 360: Unbounded anti-aliased neural radiance fields supplemental materials,'' in \emph{Proceedings of the IEEE/CVF Conference on Computer Vision and Pattern Recognition}, 2022.

\bibitem{blattmann2023stable}
A.~Blattmann, T.~Dockhorn, S.~Kulal, D.~Mendelevitch, M.~Kilian, D.~Lorenz, Y.~Levi, Z.~English, V.~Voleti, A.~Letts \emph{et~al.}, ``Stable video diffusion: Scaling latent video diffusion models to large datasets,'' \emph{arXiv preprint arXiv:2311.15127}, 2023.

\bibitem{blattmann2023align}
A.~Blattmann, R.~Rombach, H.~Ling, T.~Dockhorn, S.~W. Kim, S.~Fidler, and K.~Kreis, ``Align your latents: High-resolution video synthesis with latent diffusion models,'' in \emph{Proceedings of the IEEE/CVF Conference on Computer Vision and Pattern Recognition}, 2023, pp. 22\,563--22\,575.

\bibitem{brooks2024video}
T.~Brooks, B.~Peebles, C.~Holmes, W.~DePue, Y.~Guo, L.~Jing, D.~Schnurr, J.~Taylor, T.~Luhman, E.~Luhman \emph{et~al.}, ``Video generation models as world simulators,'' \emph{OpenAI Blog}, vol.~1, p.~8, 2024.

\bibitem{caron2021emerging}
M.~Caron, H.~Touvron, I.~Misra, H.~J{\'e}gou, J.~Mairal, P.~Bojanowski, and A.~Joulin, ``Emerging properties in self-supervised vision transformers,'' in \emph{Proceedings of the IEEE/CVF International Conference on Computer Vision}, 2021, pp. 9650--9660.

\bibitem{chai2024star}
Z.~Chai, C.~Tang, Y.~Wong, and M.~Kankanhalli, ``{STAR}: Skeleton-aware text-based {4D} avatar generation with in-network motion retargeting,'' \emph{arXiv preprint arXiv:2406.04629}, 2024.

\bibitem{chan2023generative}
E.~R. Chan, K.~Nagano, M.~A. Chan, A.~W. Bergman, J.~J. Park, A.~Levy, M.~Aittala, S.~De~Mello, T.~Karras, and G.~Wetzstein, ``Generative novel view synthesis with {3D}-aware diffusion models,'' in \emph{Proceedings of the IEEE/CVF International Conference on Computer Vision}, 2023, pp. 4217--4229.

\bibitem{pixelsplats}
D.~Charatan, S.~L. Li, A.~Tagliasacchi, and V.~Sitzmann, ``{pixelSplat}: {3D} {Gaussian} splats from image pairs for scalable generalizable {3D} reconstruction,'' in \emph{Proceedings of the IEEE/CVF Conference on Computer Vision and Pattern Recognition (CVPR)}, 2023, pp. 19\,457--19\,467.

\bibitem{chen2023videocrafter1}
H.~Chen, M.~Xia, Y.~He, Y.~Zhang, X.~Cun, S.~Yang, J.~Xing, Y.~Liu, Q.~Chen, X.~Wang \emph{et~al.}, ``{VideoCrafter1}: Open diffusion models for high-quality video generation,'' \emph{arXiv preprint arXiv:2310.19512}, 2023.

\bibitem{chen2023pixart}
J.~Chen, J.~Yu, C.~Ge, L.~Yao, E.~Xie, Y.~Wu, Z.~Wang, J.~Kwok, P.~Luo, H.~Lu \emph{et~al.}, ``{PixArt}-$\alpha$: Fast training of diffusion transformer for photorealistic text-to-image synthesis,'' \emph{arXiv preprint arXiv:2310.00426}, 2023.

\bibitem{chen2023fantasia3d}
R.~Chen, Y.~Chen, N.~Jiao, and K.~Jia, ``{Fantasia3D}: Disentangling geometry and appearance for high-quality text-to-{3D} content creation,'' in \emph{Proceedings of the IEEE/CVF International Conference on Computer Vision}, 2023, pp. 22\,246--22\,256.

\bibitem{chen2024liftimage3d}
Y.~Chen, C.~Yang, J.~Fang, X.~Zhang, L.~Xie, W.~Shen, W.~Dai, H.~Xiong, and Q.~Tian, ``{LiftImage3D}: Lifting any single image to {3D} {Gaussians} with video generation priors,'' \emph{arXiv preprint arXiv:2412.09597}, 2024.

\bibitem{chen2024mvsplat}
Y.~Chen, H.~Xu, C.~Zheng, B.~Zhuang, M.~Pollefeys, A.~Geiger, T.-J. Cham, and J.~Cai, ``{MVSplat}: Efficient {3D} {Gaussian} splatting from sparse multi-view images,'' in \emph{Proceedings of the European Conference on Computer Vision}, 2024.

\bibitem{chen2024v3d}
Z.~Chen, Y.~Wang, F.~Wang, Z.~Wang, and H.~Liu, ``{V3D}: Video diffusion models are effective {3D} generators,'' \emph{arXiv preprint arXiv:2403.06738}, 2024.

\bibitem{cheong2024boosting}
S.~Y. Cheong, D.~Ceylan, A.~Mustafa, A.~Gilbert, and C.-H.~P. Huang, ``Boosting camera motion control for video diffusion transformers,'' \emph{arXiv preprint arXiv:2410.10802}, 2024.

\bibitem{lucid}
J.~Chung, S.~Lee, H.~Nam, J.~Lee, and K.~M. Lee, ``{LucidDreamer}: Domain-free generation of {3D} {Gaussian} splatting scenes,'' \emph{arXiv preprint arXiv:2311.13384}, 2023.

\bibitem{dao2023flashattention}
T.~Dao, ``{FlashAttention-2}: Faster attention with better parallelism and work partitioning,'' \emph{arXiv preprint arXiv:2307.08691}, 2023.

\bibitem{diederik2014adam}
P.~K. Diederik and J.~Ba, ``Adam: A method for stochastic optimization,'' \emph{arXiv preprint arXiv:1412.6980}, 2017.

\bibitem{dosovitskiy2016generating}
A.~Dosovitskiy and T.~Brox, ``Generating images with perceptual similarity metrics based on deep networks,'' \emph{Advances in Neural Information Processing Systems}, vol.~29, 2016.

\bibitem{feng2024geogs3d}
Q.~Feng, Z.~Xing, Z.~Wu, and Y.-G. Jiang, ``{GeoGS3D}: Single-view {3D} reconstruction via geometric-aware diffusion model and {Gaussian} splatting,'' \emph{arXiv preprint arXiv:2403.10242}, 2024.

\bibitem{cat3d}
R.~Gao, A.~Holynski, P.~Henzler, A.~Brussee, R.~Martin~Brualla, P.~Srinivasan, J.~Barron, and B.~Poole, ``{CAT3D}: Create anything in {3D} with multi-view diffusion models,'' \emph{Advances in Neural Information Processing Systems}, vol.~37, pp. 75\,468--75\,494, 2025.

\bibitem{griffiths2014principles}
P.~Griffiths and J.~Harris, \emph{Principles of Algebraic Geometry}.\hskip 1em plus 0.5em minus 0.4em\relax John Wiley \& Sons, 2014.

\bibitem{gu2023nerfdiff}
J.~Gu, A.~Trevithick, K.-E. Lin, J.~M. Susskind, C.~Theobalt, L.~Liu, and R.~Ramamoorthi, ``{NerfDiff}: Single-image view synthesis with {NeRF}-guided distillation from {3D}-aware diffusion,'' in \emph{Proceedings of the International Conference on Machine Learning}.\hskip 1em plus 0.5em minus 0.4em\relax PMLR, 2023, pp. 11\,808--11\,826.

\bibitem{sparsenerf}
Guangcong, Z.~Chen, C.~C. Loy, and Z.~Liu, ``{SparseNeRF}: Distilling depth ranking for few-shot novel view synthesis,'' in \emph{Proceedings of the IEEE/CVF International Conference on Computer Vision (ICCV)}, 2023.

\bibitem{AnimateDiff}
Y.~Guo, C.~Yang, A.~Rao, Y.~Wang, Y.~Qiao, D.~Lin, and B.~Dai, ``{AnimateDiff}: Animate your personalized text-to-image diffusion models without specific tuning,'' in \emph{Proceedings of the International Conference on Learning Representations}, 2024.

\bibitem{han2024vfusion3d}
J.~Han, F.~Kokkinos, and P.~Torr, ``{VFusion3D}: Learning scalable {3D} generative models from video diffusion models,'' in \emph{Proceedings of the European Conference on Computer Vision}.\hskip 1em plus 0.5em minus 0.4em\relax Springer, 2024, pp. 333--350.

\bibitem{he2024cameractrl}
H.~He, Y.~Xu, Y.~Guo, G.~Wetzstein, B.~Dai, H.~Li, and C.~Yang, ``{CameraCtrl}: Enabling camera control for text-to-video generation,'' \emph{arXiv preprint arXiv:2404.02101}, 2024.

\bibitem{he2025cameractrl}
H.~He, C.~Yang, S.~Lin, Y.~Xu, M.~Wei, L.~Gui, Q.~Zhao, G.~Wetzstein, L.~Jiang, and H.~Li, ``{CameraCtrl II}: Dynamic scene exploration via camera-controlled video diffusion models,'' \emph{arXiv preprint arXiv:2503.10592}, 2025.

\bibitem{he2022masked}
K.~He, X.~Chen, S.~Xie, Y.~Li, P.~Doll{\'a}r, and R.~Girshick, ``Masked autoencoders are scalable vision learners,'' in \emph{Proceedings of the IEEE/CVF Conference on Computer Vision and Pattern Recognition}, 2022, pp. 16\,000--16\,009.

\bibitem{he2024gvgen}
X.~He, J.~Chen, S.~Peng, D.~Huang, Y.~Li, X.~Huang, C.~Yuan, W.~Ouyang, and T.~He, ``{GVGEN}: Text-to-{3D} generation with volumetric representation,'' in \emph{Proceedings of the European Conference on Computer Vision}.\hskip 1em plus 0.5em minus 0.4em\relax Springer, 2024, pp. 463--479.

\bibitem{heusel2017gans}
M.~Heusel, H.~Ramsauer, T.~Unterthiner, B.~Nessler, and S.~Hochreiter, ``{GANS} trained by a two time-scale update rule converge to a local {Nash} equilibrium,'' \emph{Advances in Neural Information Processing Systems}, vol.~30, 2017.

\bibitem{ho2020denoising}
J.~Ho, A.~Jain, and P.~Abbeel, ``Denoising diffusion probabilistic models,'' \emph{Advances in Neural Information Processing Systems}, vol.~33, pp. 6840--6851, 2020.

\bibitem{hollein2024viewdiff}
L.~H{\"o}llein, A.~Bo{\v{z}}i{\v{c}}, N.~M{\"u}ller, D.~Novotny, H.-Y. Tseng, C.~Richardt, M.~Zollh{\"o}fer, and M.~Nie{\ss}ner, ``{ViewDiff}: {3D}-consistent image generation with text-to-image models,'' in \emph{Proceedings of the IEEE/CVF Conference on Computer Vision and Pattern Recognition}, 2024, pp. 5043--5052.

\bibitem{lrm}
Y.~Hong, K.~Zhang, J.~Gu, S.~Bi, Y.~Zhou, D.~Liu, F.~Liu, K.~Sunkavalli, T.~Bui, and H.~Tan, ``{LRM}: Large reconstruction model for single image to {3D},'' \emph{arXiv preprint arXiv:2311.04400}, 2023.

\bibitem{hou2024training}
C.~Hou, G.~Wei, Y.~Zeng, and Z.~Chen, ``Training-free camera control for video generation,'' \emph{arXiv preprint arXiv:2406.10126}, 2024.

\bibitem{hu2021lora}
E.~J. Hu, Y.~Shen, P.~Wallis, Z.~Allen-Zhu, Y.~Li, S.~Wang, L.~Wang, and W.~Chen, ``{LoRA}: Low-rank adaptation of large language models,'' \emph{arXiv preprint arXiv:2106.09685}, 2021.

\bibitem{hu2024motionmaster}
T.~Hu, J.~Zhang, R.~Yi, Y.~Wang, H.~Huang, J.~Weng, Y.~Wang, and L.~Ma, ``{MotionMaster}: Training-free camera motion transfer for video generation,'' \emph{arXiv preprint arXiv:2404.15789}, 2024.

\bibitem{isola2017image}
P.~Isola, J.-Y. Zhu, T.~Zhou, and A.~A. Efros, ``Image-to-image translation with conditional adversarial networks,'' in \emph{Proceedings of the IEEE Conference on Computer Vision and Pattern Recognition}, 2017, pp. 1125--1134.

\bibitem{DietNeRF}
A.~Jain, M.~Tancik, and P.~Abbeel, ``Putting {NeRF} on a diet: Semantically consistent few-shot view synthesis,'' in \emph{CVPR}, 2021, pp. 5885--5894.

\bibitem{jiang2024animate3d}
Y.~Jiang, C.~Yu, C.~Cao, F.~Wang, W.~Hu, and J.~Gao, ``{Animate3D}: Animating any {3D} model with multi-view video diffusion,'' \emph{arXiv preprint arXiv:2407.11398}, 2024.

\bibitem{jin2024lvsm}
H.~Jin, H.~Jiang, H.~Tan, K.~Zhang, S.~Bi, T.~Zhang, F.~Luan, N.~Snavely, and Z.~Xu, ``{LVSM}: A large view synthesis model with minimal {3D} inductive bias,'' \emph{arXiv preprint arXiv:2410.17242}, 2024.

\bibitem{jin2025flovd}
W.~Jin, Q.~Dai, C.~Luo, S.-H. Baek, and S.~Cho, ``{FloVD}: Optical flow meets video diffusion model for enhanced camera-controlled video synthesis,'' \emph{arXiv preprint arXiv:2502.08244}, 2025.

\bibitem{3dgs}
B.~Kerbl, G.~Kopanas, T.~Leimk{\"u}hler, and G.~Drettakis, ``{3D} {Gaussian} splatting for real-time radiance field rendering,'' \emph{ACM Transactions on Graphics}, vol.~42, no.~4, 2023.

\bibitem{kim2023neuralfield}
S.~W. Kim, B.~Brown, K.~Yin, K.~Kreis, K.~Schwarz, D.~Li, R.~Rombach, A.~Torralba, and S.~Fidler, ``{NeuralField-LDM}: Scene generation with hierarchical latent diffusion models,'' in \emph{Proceedings of the IEEE/CVF Conference on Computer Vision and Pattern Recognition}, 2023, pp. 8496--8506.

\bibitem{Knapitsch2017}
A.~Knapitsch, J.~Park, Q.-Y. Zhou, and V.~Koltun, ``Tanks and temples: Benchmarking large-scale scene reconstruction,'' \emph{ACM Transactions on Graphics}, vol.~36, no.~4, 2017.

\bibitem{kuang2024collaborative}
Z.~Kuang, S.~Cai, H.~He, Y.~Xu, H.~Li, L.~J. Guibas, and G.~Wetzstein, ``Collaborative video diffusion: Consistent multi-video generation with camera control,'' \emph{Advances in Neural Information Processing Systems}, vol.~37, pp. 16\,240--16\,271, 2024.

\bibitem{flux_repo}
\BIBentryALTinterwordspacing
B.~F. Labs, ``Flux: Decentralized computation framework,'' 2023, accessed: 2024-11-14. [Online]. Available: \url{https://github.com/black-forest-labs/flux}
\BIBentrySTDinterwordspacing

\bibitem{li2024vivid}
B.~Li, C.~Zheng, W.~Zhu, J.~Mai, B.~Zhang, P.~Wonka, and B.~Ghanem, ``{Vivid-ZOO}: Multi-view video generation with diffusion model,'' \emph{Advances in Neural Information Processing Systems}, vol.~37, pp. 62\,189--62\,222, 2024.

\bibitem{li2023instant3d}
J.~Li, H.~Tan, K.~Zhang, Z.~Xu, F.~Luan, Y.~Xu, Y.~Hong, K.~Sunkavalli, G.~Shakhnarovich, and S.~Bi, ``{Instant3D}: Fast text-to-{3D} with sparse-view generation and large reconstruction model,'' \emph{arXiv preprint arXiv:2311.06214}, 2023.

\bibitem{li2025realcam}
T.~Li, G.~Zheng, R.~Jiang, T.~Wu, Y.~Lu, Y.~Lin, X.~Li \emph{et~al.}, ``{RealCam-I2V}: Real-world image-to-video generation with interactive complex camera control,'' \emph{arXiv preprint arXiv:2502.10059}, 2025.

\bibitem{liang2022self}
H.~Liang, N.~Quader, Z.~Chi, L.~Chen, P.~Dai, J.~Lu, and Y.~Wang, ``Self-supervised spatiotemporal representation learning by exploiting video continuity,'' in \emph{Proceedings of the AAAI conference on artificial intelligence}, vol.~36, no.~2, 2022, pp. 1564--1573.

\bibitem{liang2024diffusion4d}
H.~Liang, Y.~Yin, D.~Xu, H.~Liang, Z.~Wang, K.~N. Plataniotis, Y.~Zhao, and Y.~Wei, ``Diffusion4d: Fast spatial-temporal consistent 4d generation via video diffusion models,'' \emph{Advances in Neural Information Processing Systems}, vol.~37, pp. 110\,854--110\,875, 2024.

\bibitem{liang2024feed}
H.~Liang, J.~Ren, A.~Mirzaei, A.~Torralba, Z.~Liu, I.~Gilitschenski, S.~Fidler, C.~Oztireli, H.~Ling, Z.~Gojcic \emph{et~al.}, ``Feed-forward bullet-time reconstruction of dynamic scenes from monocular videos,'' \emph{arXiv preprint arXiv:2412.03526}, 2024.

\bibitem{ling2024dl3dv}
L.~Ling, Y.~Sheng, Z.~Tu, W.~Zhao, C.~Xin, K.~Wan, L.~Yu, Q.~Guo, Z.~Yu, Y.~Lu \emph{et~al.}, ``{Dl3dv-10k}: A large-scale scene dataset for deep learning-based {3D} vision,'' in \emph{Proceedings of the IEEE/CVF Conference on Computer Vision and Pattern Recognition}, 2024, pp. 22\,160--22\,169.

\bibitem{liu2021infinite}
A.~Liu, R.~Tucker, V.~Jampani, A.~Makadia, N.~Snavely, and A.~Kanazawa, ``Infinite nature: Perpetual view generation of natural scenes from a single image,'' in \emph{Proceedings of the IEEE/CVF International Conference on Computer Vision}, 2021, pp. 14\,458--14\,467.

\bibitem{reconx}
F.~Liu, W.~Sun, H.~Wang, Y.~Wang, H.~Sun, J.~Ye, J.~Zhang, and Y.~Duan, ``{ReconX}: Reconstruct any scene from sparse views with video diffusion model,'' \emph{arXiv preprint arXiv:2408.16767}, 2024.

\bibitem{liu2024one}
M.~Liu, R.~Shi, L.~Chen, Z.~Zhang, C.~Xu, X.~Wei, H.~Chen, C.~Zeng, J.~Gu, and H.~Su, ``One-2-3-45++: Fast single image to {3D} objects with consistent multi-view generation and {3D} diffusion,'' in \emph{Proceedings of the IEEE/CVF Conference on Computer Vision and Pattern Recognition}, 2024, pp. 10\,072--10\,083.

\bibitem{zero123}
R.~Liu, R.~Wu, B.~V. Hoorick, P.~Tokmakov, S.~Zakharov, and C.~Vondrick, ``Zero-1-to-3: Zero-shot one image to {3D} object,'' \emph{arXiv preprint arXiv:2303.11328}, 2023.

\bibitem{liu2024sora}
Y.~Liu, K.~Zhang, Y.~Li, Z.~Yan, C.~Gao, R.~Chen, Z.~Yuan, Y.~Huang, H.~Sun, J.~Gao \emph{et~al.}, ``Sora: A review on background, technology, limitations, and opportunities of large vision models,'' \emph{arXiv preprint arXiv:2402.17177}, 2024.

\bibitem{long2024wonder3d}
X.~Long, Y.-C. Guo, C.~Lin, Y.~Liu, Z.~Dou, L.~Liu, Y.~Ma, S.-H. Zhang, M.~Habermann, C.~Theobalt \emph{et~al.}, ``{Wonder3D}: Single image to {3D} using cross-domain diffusion,'' in \emph{Proceedings of the IEEE/CVF Conference on Computer Vision and Pattern Recognition}, 2024, pp. 9970--9980.

\bibitem{lu2024genex}
T.~Lu, T.~Shu, J.~Xiao, L.~Ye, J.~Wang, C.~Peng, C.~Wei, D.~Khashabi, R.~Chellappa, A.~Yuille \emph{et~al.}, ``{GenEx}: Generating an explorable world,'' \emph{arXiv preprint arXiv:2412.09624}, 2024.

\bibitem{TrackNeRF}
J.~Mai, W.~Zhu, S.~Rojas, J.~Zarzar, A.~Hamdi, G.~Qian, B.~Li, S.~Giancola, and B.~Ghanem, ``{TrackNeRF}: Bundle adjusting {NeRF} from sparse and noisy views via feature tracks,'' in \emph{Proceedings of the IEEE/CVF Conference on Computer Vision and Pattern Recognition}, vol. abs/2408.10739, 2024.

\bibitem{menapace2024snap}
W.~Menapace, A.~Siarohin, I.~Skorokhodov, E.~Deyneka, T.-S. Chen, A.~Kag, Y.~Fang, A.~Stoliar, E.~Ricci, J.~Ren \emph{et~al.}, ``Snap video: Scaled spatiotemporal transformers for text-to-video synthesis,'' in \emph{Proceedings of the IEEE/CVF Conference on Computer Vision and Pattern Recognition}, 2024, pp. 7038--7048.

\bibitem{micikevicius2017mixed}
P.~Micikevicius, S.~Narang, J.~Alben, G.~Diamos, E.~Elsen, D.~Garcia, B.~Ginsburg, M.~Houston, O.~Kuchaiev, G.~Venkatesh \emph{et~al.}, ``Mixed precision training,'' \emph{arXiv preprint arXiv:1710.03740}, 2017.

\bibitem{nerf}
B.~Mildenhall, P.~P. Srinivasan, M.~Tancik, J.~T. Barron, R.~Ramamoorthi, and R.~Ng, ``{NeRF}: Representing scenes as neural radiance fields for view synthesis,'' \emph{Communications of the ACM}, vol.~65, no.~1, pp. 99--106, 2021.

\bibitem{peebles2023scalable}
W.~Peebles and S.~Xie, ``Scalable diffusion models with transformers,'' in \emph{Proceedings of the IEEE/CVF International Conference on Computer Vision}, 2023, pp. 4195--4205.

\bibitem{podell2023sdxl}
D.~Podell, Z.~English, K.~Lacey, A.~Blattmann, T.~Dockhorn, J.~M{\"u}ller, J.~Penna, and R.~Rombach, ``{SDXL}: Improving latent diffusion models for high-resolution image synthesis,'' \emph{arXiv preprint arXiv:2307.01952}, 2023.

\bibitem{DreamFusion}
B.~Poole, A.~Jain, J.~T. Barron, and B.~Mildenhall, ``{DreamFusion}: Text-to-{3D} using {2D} diffusion,'' in \emph{Proceedings of the International Conference on Learning Representations}, 2022.

\bibitem{Magic123}
G.~Qian, J.~Mai, A.~Hamdi, J.~Ren, A.~Siarohin, B.~Li, H.~Lee, I.~Skorokhodov, P.~Wonka, S.~Tulyakov, and B.~Ghanem, ``Magic123: One image to high-quality {3D} object generation using both {2D} and {3D} diffusion priors,'' in \emph{Proceedings of the International Conference on Learning Representations}, 2024.

\bibitem{ren2024l4gm}
J.~Ren, K.~Xie, A.~Mirzaei, H.~Liang, X.~Zeng, K.~Kreis, Z.~Liu, A.~Torralba, S.~Fidler, S.~W. Kim \emph{et~al.}, ``{L4GM}: Large {4D} {Gaussian} reconstruction model,'' \emph{arXiv preprint arXiv:2406.10324}, 2024.

\bibitem{latentdiffusion}
R.~Rombach, A.~Blattmann, D.~Lorenz, P.~Esser, and B.~Ommer, ``High-resolution image synthesis with latent diffusion models,'' in \emph{Proceedings of the IEEE/CVF Conference on Computer Vision and Pattern Recognition}, 2022, pp. 10\,684--10\,695.

\bibitem{zeronvs}
K.~Sargent, Z.~Li, T.~Shah, C.~Herrmann, H.-X. Yu, Y.~Zhang, E.~R. Chan, D.~Lagun, L.~Fei-Fei, D.~Sun, and J.~Wu, ``{ZeroNVS}: Zero-shot 360-degree view synthesis from a single image,'' \emph{arXiv preprint arXiv:2310.17994}, 2024.

\bibitem{schoenberger2016sfm}
J.~L. Sch\"{o}nberger and J.-M. Frahm, ``Structure-from-motion revisited,'' in \emph{Proceedings of the IEEE Conference on Computer Vision and Pattern Recognition (CVPR)}, 2016.

\bibitem{schoenberger2016mvs}
J.~L. Sch\"{o}nberger, E.~Zheng, M.~Pollefeys, and J.-M. Frahm, ``Pixelwise view selection for unstructured multi-view stereo,'' in \emph{Proceedings of the European Conference on Computer Vision (ECCV)}, 2016.

\bibitem{schwarz2025generative}
K.~Schwarz, N.~Mueller, and P.~Kontschieder, ``Generative {Gaussian} splatting: Generating {3D} scenes with video diffusion priors,'' \emph{arXiv preprint arXiv:2503.13272}, 2025.

\bibitem{shi2023mvdream}
Y.~Shi, P.~Wang, J.~Ye, M.~Long, K.~Li, and X.~Yang, ``{MVDream}: Multi-view diffusion for {3D} generation,'' \emph{arXiv preprint arXiv:2308.16512}, 2023.

\bibitem{realdreamer}
J.~Shriram, A.~Trevithick, L.~Liu, and R.~Ramamoorthi, ``{RealmDreamer}: Text-driven {3D} scene generation with inpainting and depth diffusion,'' \emph{arXiv preprint arXiv:2404.07199}, 2024.

\bibitem{simonyan2014very}
K.~Simonyan, ``Very deep convolutional networks for large-scale image recognition,'' \emph{arXiv preprint arXiv:1409.1556}, 2014.

\bibitem{song2020score}
Y.~Song, J.~Sohl-Dickstein, D.~P. Kingma, A.~Kumar, S.~Ermon, and B.~Poole, ``Score-based generative modeling through stochastic differential equations,'' \emph{arXiv preprint arXiv:2011.13456}, 2020.

\bibitem{sun2024dimensionx}
W.~Sun, S.~Chen, F.~Liu, Z.~Chen, Y.~Duan, J.~Zhang, and Y.~Wang, ``{DimensionX}: Create any {3D} and {4D} scenes from a single image with controllable video diffusion,'' \emph{arXiv preprint arXiv:2411.04928}, 2024.

\bibitem{szymanowicz2024flash3d}
S.~Szymanowicz, E.~Insafutdinov, C.~Zheng, D.~Campbell, J.~F. Henriques, C.~Rupprecht, and A.~Vedaldi, ``{Flash3D}: Feed-forward generalisable {3D} scene reconstruction from a single image,'' \emph{arXiv preprint arXiv:2406.04343}, 2024.

\bibitem{szymanowicz2024splatter}
S.~Szymanowicz, C.~Rupprecht, and A.~Vedaldi, ``Splatter image: Ultra-fast single-view {3D} reconstruction,'' in \emph{Proceedings of the IEEE/CVF Conference on Computer Vision and Pattern Recognition}, 2024, pp. 10\,208--10\,217.

\bibitem{szymanowicz2025bolt3d}
S.~Szymanowicz, J.~Y. Zhang, P.~Srinivasan, R.~Gao, A.~Brussee, A.~Holynski, R.~Martin-Brualla, J.~T. Barron, and P.~Henzler, ``{Bolt3D}: Generating {3D} scenes in seconds,'' \emph{arXiv preprint arXiv:2503.14445}, 2025.

\bibitem{lgm}
J.~Tang, Z.~Chen, X.~Chen, T.~Wang, G.~Zeng, and Z.~Liu, ``{LGM}: Large multi-view {Gaussian} model for high-resolution {3D} content creation,'' \emph{arXiv preprint arXiv:2402.05054}, 2024.

\bibitem{tochilkin2024triposr}
D.~Tochilkin, D.~Pankratz, Z.~Liu, Z.~Huang, A.~Letts, Y.~Li, D.~Liang, C.~Laforte, V.~Jampani, and Y.-P. Cao, ``{TripoSR}: Fast {3D} object reconstruction from a single image,'' \emph{arXiv preprint arXiv:2403.02151}, 2024.

\bibitem{fvd}
T.~Unterthiner, S.~van Steenkiste, K.~Kurach, R.~Marinier, M.~Michalski, and S.~Gelly, ``{FVD}: A new metric for video generation,'' in \emph{Proceedings of the ICLR Workshop on Deep Generative Models for Highly Structured Data}, 2019.

\bibitem{voleti2024sv3d}
V.~Voleti, C.-H. Yao, M.~Boss, A.~Letts, D.~Pankratz, D.~Tochilkin, C.~Laforte, R.~Rombach, and V.~Jampani, ``{SV3D}: Novel multi-view synthesis and {3D} generation from a single image using latent video diffusion,'' in \emph{Proceedings of the European Conference on Computer Vision}.\hskip 1em plus 0.5em minus 0.4em\relax Springer, 2024, pp. 439--457.

\bibitem{wang2024vistadream}
H.~Wang, Y.~Liu, Z.~Liu, W.~Wang, Z.~Dong, and B.~Yang, ``{VistaDream}: Sampling multiview consistent images for single-view scene reconstruction,'' \emph{arXiv preprint arXiv:2410.16892}, 2024.

\bibitem{wang2024cpa}
Y.~Wang, J.~Zhang, P.~Jiang, H.~Zhang, J.~Chen, and B.~Li, ``{CPA}: Camera-pose-awareness diffusion transformer for video generation,'' \emph{arXiv preprint arXiv:2412.01429}, 2024.

\bibitem{wang2023prolificdreamer}
Z.~Wang, C.~Lu, Y.~Wang, F.~Bao, C.~Li, H.~Su, and J.~Zhu, ``{ProlificDreamer}: High-fidelity and diverse text-to-{3D} generation with variational score distillation,'' \emph{Advances in Neural Information Processing Systems}, vol.~36, pp. 8406--8441, 2023.

\bibitem{ssim}
Z.~Wang, A.~C. Bovik, H.~R. Sheikh, and E.~P. Simoncelli, ``Image quality assessment: From error visibility to structural similarity,'' \emph{IEEE Transactions on Image Processing}, vol.~13, no.~4, pp. 600--612, 2004.

\bibitem{wang2024motionctrl}
Z.~Wang, Z.~Yuan, X.~Wang, Y.~Li, T.~Chen, M.~Xia, P.~Luo, and Y.~Shan, ``{MotionCtrl}: A unified and flexible motion controller for video generation,'' in \emph{Proceedings of ACM SIGGRAPH 2024}, 2024, pp. 1--11.

\bibitem{reconfusion}
R.~Wu, B.~Mildenhall, P.~Henzler, K.~Park, R.~Gao, D.~Watson, P.~P. Srinivasan, D.~Verbin, J.~T. Barron, B.~Poole, and A.~Holynski, ``{ReconFusion}: {3D} reconstruction with diffusion priors,'' \emph{arXiv preprint arXiv:2312.02981}, 2023.

\bibitem{xiao20243dtrajmaster}
F.~Xiao, X.~Liu, X.~Wang, S.~Peng, M.~Xia, X.~Shi, Z.~Yuan, P.~Wan, D.~Zhang, and D.~Lin, ``{3DTrajMaster}: Mastering {3D} trajectory for multi-entity motion in video generation,'' in \emph{Proceedings of the International Conference on Learning Representations}, 2024.

\bibitem{xiao2024trajectory}
Z.~Xiao, W.~Ouyang, Y.~Zhou, S.~Yang, L.~Yang, J.~Si, and X.~Pan, ``Trajectory attention for fine-grained video motion control,'' \emph{arXiv preprint arXiv:2411.19324}, 2024.

\bibitem{xie2024sv4d}
Y.~Xie, C.-H. Yao, V.~Voleti, H.~Jiang, and V.~Jampani, ``{SV4D}: Dynamic {3D} content generation with multi-frame and multi-view consistency,'' \emph{arXiv preprint arXiv:2407.17470}, 2024.

\bibitem{xu2024comp4d}
D.~Xu, H.~Liang, N.~P. Bhatt, H.~Hu, H.~Liang, K.~N. Plataniotis, and Z.~Wang, ``{Comp4D}: {LLM}-guided compositional {4D} scene generation,'' \emph{arXiv preprint arXiv:2403.16993}, 2024.

\bibitem{xu2024camco}
D.~Xu, W.~Nie, C.~Liu, S.~Liu, J.~Kautz, Z.~Wang, and A.~Vahdat, ``{CamCo}: Camera-controllable {3D}-consistent image-to-video generation,'' \emph{arXiv preprint arXiv:2406.02509}, 2024.

\bibitem{xu2024agg}
D.~Xu, Y.~Yuan, M.~Mardani, S.~Liu, J.~Song, Z.~Wang, and A.~Vahdat, ``{AGG}: Amortized generative {3D} {Gaussians} for single image to {3D},'' \emph{arXiv preprint arXiv:2401.04099}, 2024.

\bibitem{xu2024controlrm}
H.~Xu, W.~Chen, Z.~Zhou, F.~Xiao, B.~Sun, M.~Z. Shou, and W.~Kang, ``{ControLRM}: Fast and controllable {3D} generation via large reconstruction model,'' \emph{arXiv preprint arXiv:2410.09592}, 2024.

\bibitem{xu2024grm}
Y.~Xu, Z.~Shi, W.~Yifan, H.~Chen, C.~Yang, S.~Peng, Y.~Shen, and G.~Wetzstein, ``{GRM}: Large {Gaussian} reconstruction model for efficient {3D} reconstruction and generation,'' in \emph{Proceedings of the European Conference on Computer Vision}.\hskip 1em plus 0.5em minus 0.4em\relax Springer, 2024, pp. 1--20.

\bibitem{yang2024diffusion}
Z.~Yang, Z.~Pan, C.~Gu, and L.~Zhang, ``Diffusion$^2$: Dynamic {3D} content generation via score composition of orthogonal diffusion models,'' \emph{arXiv preprint arXiv:2404.02148}, 2024.

\bibitem{cogvideo}
Z.~Yang, J.~Teng, W.~Zheng, M.~Ding, S.~Huang, J.~Xu, Y.~Yang, W.~Hong, X.~Zhang, G.~Feng, D.~Yin, X.~Gu, Y.~Zhang, W.~Wang, Y.~Cheng, T.~Liu, B.~Xu, Y.~Dong, and J.~Tang, ``{CogVideoX}: Text-to-video diffusion models with an expert transformer,'' \emph{arXiv preprint arXiv:2408.06072}, 2024.

\bibitem{pixelnerf}
A.~Yu, V.~Ye, M.~Tancik, and A.~Kanazawa, ``{pixelNeRF}: Neural radiance fields from one or few images,'' \emph{arXiv preprint arXiv:2012.02190}, 2021.

\bibitem{yu2024wonderworld}
H.-X. Yu, H.~Duan, C.~Herrmann, W.~T. Freeman, and J.~Wu, ``{WonderWorld}: Interactive {3D} scene generation from a single image,'' \emph{arXiv preprint arXiv:2406.09394}, 2024.

\bibitem{yu2024wonderjourney}
H.-X. Yu, H.~Duan, J.~Hur, K.~Sargent, M.~Rubinstein, W.~T. Freeman, F.~Cole, D.~Sun, N.~Snavely, J.~Wu \emph{et~al.}, ``{WonderJourney}: Going from anywhere to everywhere,'' in \emph{Proceedings of the IEEE/CVF Conference on Computer Vision and Pattern Recognition}, 2024, pp. 6658--6667.

\bibitem{yu2021vector}
J.~Yu, X.~Li, J.~Y. Koh, H.~Zhang, R.~Pang, J.~Qin, A.~Ku, Y.~Xu, J.~Baldridge, and Y.~Wu, ``Vector-quantized image modeling with improved {VQGAN},'' \emph{arXiv preprint arXiv:2110.04627}, 2021.

\bibitem{viewcrafter}
W.~Yu, J.~Xing, L.~Yuan, W.~Hu, X.~Li, Z.~Huang, X.~Gao, T.-T. Wong, Y.~Shan, and Y.~Tian, ``{ViewCrafter}: Taming video diffusion models for high-fidelity novel view synthesis,'' \emph{arXiv preprint arXiv:2409.02048}, 2024.

\bibitem{yuan2024generative}
Y.~Yuan, X.~Wang, Y.~Sheng, P.~Chennuri, X.~Zhang, and S.~Chan, ``Generative photography: Scene-consistent camera control for realistic text-to-image synthesis,'' \emph{arXiv preprint arXiv:2412.02168}, 2024.

\bibitem{gslrm}
K.~Zhang, S.~Bi, H.~Tan, Y.~Xiangli, N.~Zhao, K.~Sunkavalli, and Z.~Xu, ``{GS-LRM}: Large reconstruction model for {3D} {Gaussian} splatting,'' \emph{arXiv preprint arXiv:2404.19702}, 2024.

\bibitem{zhang2023adding}
L.~Zhang, A.~Rao, and M.~Agrawala, ``Adding conditional control to text-to-image diffusion models,'' in \emph{Proceedings of the IEEE/CVF International Conference on Computer Vision}, 2023, pp. 3836--3847.

\bibitem{zhang2018unreasonable}
R.~Zhang, P.~Isola, A.~A. Efros, E.~Shechtman, and O.~Wang, ``The unreasonable effectiveness of deep features as a perceptual metric,'' in \emph{Proceedings of the IEEE Conference on Computer Vision and Pattern Recognition}, 2018, pp. 586--595.

\bibitem{zhao2024genxd}
Y.~Zhao, C.-C. Lin, K.~Lin, Z.~Yan, L.~Li, Z.~Yang, J.~Wang, G.~H. Lee, and L.~Wang, ``{GenXD}: Generating any {3D} and {4D} scenes,'' \emph{arXiv preprint arXiv:2411.02319}, 2024.

\bibitem{zhou2018stereo}
T.~Zhou, R.~Tucker, J.~Flynn, G.~Fyffe, and N.~Snavely, ``Stereo magnification: Learning view synthesis using multiplane images,'' \emph{arXiv preprint arXiv:1805.09817}, 2018.

\bibitem{ziwen2024long}
C.~Ziwen, H.~Tan, K.~Zhang, S.~Bi, F.~Luan, Y.~Hong, L.~Fuxin, and Z.~Xu, ``{Long-LRM}: Long-sequence large reconstruction model for wide-coverage {Gaussian} splats,'' \emph{arXiv preprint arXiv:2410.12781}, 2024.

\end{thebibliography}

\clearpage
 
\appendix

\setcounter{page}{1}
\setcounter{section}{0}
\setcounter{figure}{0}
\setcounter{table}{0}
\renewcommand{\thefigure}{A\arabic{figure}}
\renewcommand{\thetable}{A\arabic{table}}
\renewcommand{\thesection}{\Alph{section}}
\renewcommand \thepart{}
\renewcommand \partname{}
\renewcommand{\thesection}{\Alph{section}}

\twocolumn[{
\renewcommand\twocolumn[1][]{#1}
\maketitlesupplementary

\vspace{-2cm}
\hypersetup{linkcolor=black}
\doparttoc % Tell to minitoc to generate a toc for the parts
\faketableofcontents % Run a fake tableofcontents command for the partocs
\part{} % Start the document part
\parttoc % Insert the document TOC

\normalsize
\captionsetup{type=table}
\vspace{-0.2cm}
\caption{\textbf{Mathematical notation used in the paper.}}
\label{Tab:notation}
\begin{tabular}{lll}
\toprule
\textbf{Expression} & \textbf{Specification} & \textbf{Explanation} \\
\midrule
\textit{\textbf{Commonly Used}} & \multicolumn{2}{c}{}  \\
$x$ & $x\in R^{T\times H\times W\times 3}$ & source video clip \\
$s$ & -- & stride in sampling clip $x$ from source video \\
% $T$,$H$,$W$ & 49,480,720 & temporal length, height, width of $x$ \\
$z$ & $z \in R^{t\times h\times w\times c}$ & video latent embedded from $x$ \\
$\mathcal{E}$ & -- & encoder from 3D-VAE \\
$r_s$ & $r_s=\frac{H}{h}=\frac{W}{w}$ & spatial compression rate \\[3pt]
$r_t$ & $r_t=\frac{T}{t}$ & temporal compression rate \\[3pt]
$p$ & $p \in R^{T\times H\times W\times 6}$ & Plücker embedding of cameras of video clip $x$ \\
\midrule
\textit{\textbf{Diffusion Related}} & \multicolumn{2}{c}{}  \\
$\tau$ & -- & diffusion time step \\
$\alpha_{\tau}, \sigma_{\tau}$ & - & diffusion noise scheduler parameters \\
$\epsilon$ & $\epsilon \in R^{t\times h\times w\times c}$ & random noise such that $\epsilon \sim \mathcal{N}(0,I)$ \\
$z_\tau$ & $z_\tau \in R^{t\times h\times w\times c}$ & noisy video latent \\
$D_{\theta}$ & -- & diffusion model parameterized by $\theta$ \\
$y$ & -- & conditional signal, usually textual embedding \\
$o_v$ & $o_v\in R^{N_v\times d_v}$ & visual tokens as a sequence in the diffusion model \\
$o_\text{ctrl}$, $o_\text{lora}$ & $o_\text{ctrl}$, $o_\text{lora}$ $\in R^{N_v\times d_v}$ & camera tokens as a sequence in the diffusion model \\
$N$ & -- & number of transformer blocks in the ControlNet branch \\
\midrule
\textit{\textbf{Reconstruction Related}} & \multicolumn{2}{c}{}  \\
$p_l$ & -- & spatial patch size applied to $z$ in the LaLRM \\
$o_l$ & $o_l\in R^{N_l\times d_l}$ & visual latent tokens as a sequence in the LaLRM \\
$N_l$ & $N_l=t\cdot \frac{h}{p_l}\cdot \frac{w}{p_l}$ & number of visual latent tokens in the LaLRM \\[5pt]
$o_\text{p}$ & $o_\text{p}$ $\in R^{N_l\times d_l}$ & camera tokens as a sequence in the LaLRM \\
$V$ & -- & number of supervision views in the LaLRM \\
$G$ & $G\in R^{(T\times H\times W)\times 12}$ & Gaussian feature map in the LaLRM \\
\bottomrule
\end{tabular}
}]

\newpage

\begin{figure*}
  \centering
  % \vspace{-0.2cm}
  \includegraphics[width=1.0\linewidth]{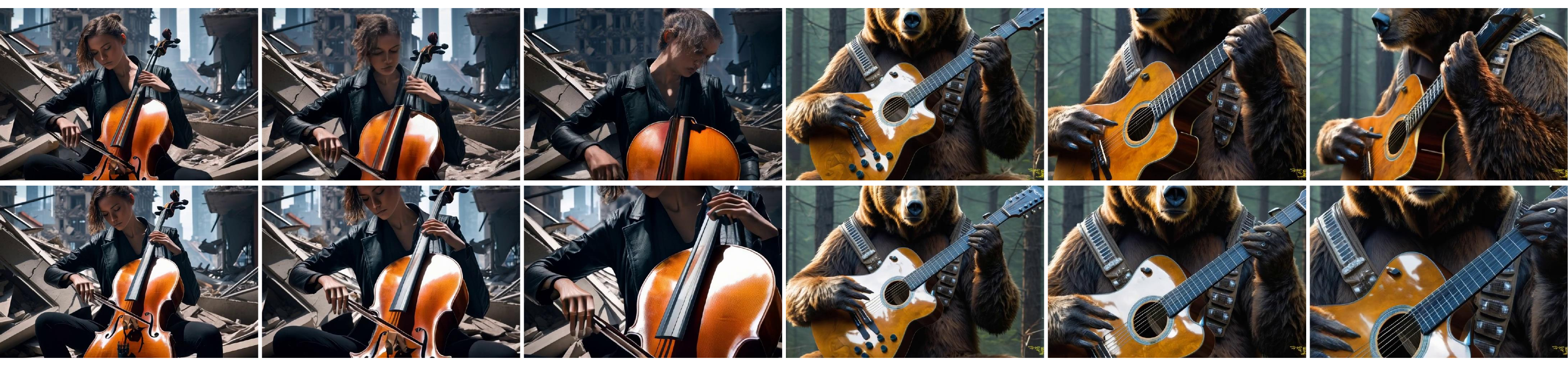}
  \vspace{-0.5cm}
  \caption{\textbf{Comparison of video generations} between the source model (\emph{top row}) and the model fine-tuned on static-scene datasets with LoRA modules (\emph{bottom row}). The results demonstrate that fine-tuning the model on static-scene datasets equipped with LoRA produces significantly more static scenes.}
  \vspace{-0.2cm}
  \label{fig:supp1}
\end{figure*}

\begin{figure*}
  \centering
  \includegraphics[width=1.0\linewidth]{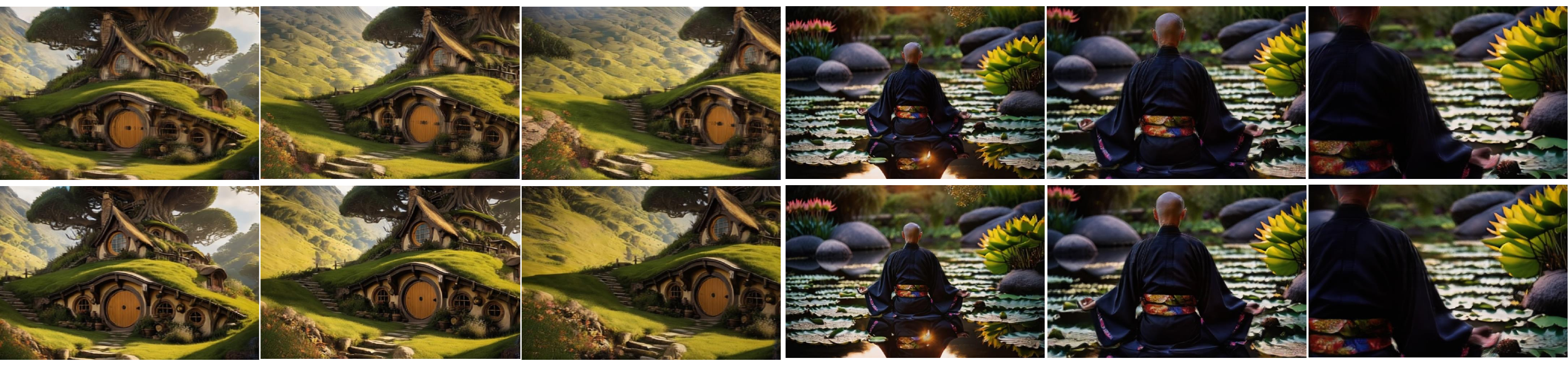}
  \vspace{-0.4cm}
  \caption{\textbf{Comparison of 3D rendering performance} between latent reconstruction models fine-tuned \textit{without} in-the-wild dataset (\emph{upper row}) and \textit{with} in-the-wild dataset (\emph{lower row}). Involving in-the-wild datasets during fine-tuning improves the generalization capability.}
  % \vspace{-0.3cm}
  \label{fig:supp2}
\end{figure*}

\section{Additional Analysis of Controllable Video Generation}
\label{sec:abl_video}

This section presents further ablation analysis of our camera-conditioned video generation pipeline.

\subsection{Effect of LoRA on Static Scene Generation and Camera Controllability}

We employ LoRA fine-tuning in the main branch of our camera-guided video generation model.
LoRA is advantageous for its compatibility with pre-trained models, as it introduces extensions without altering the original model weights.
The lightweight module offers a cost-effective approach to fine-tuning heavy models, requiring minimal computational resources and reducing the risk of overfitting to customized datasets.
Our framework takes advantage of LoRA to enhance static scene generation and camera controllability.

To evaluate the impact on \textit{static scene generation}, we fine-tuned the I2V source model using LoRA on customized datasets dominated by static scenes, including RealEstate10K (RE10K), ACID, and DL3DV. 
In this scenario, no pose control was applied. 
We compared the performance of the fine-tuned model against the source model on 20 in-the-wild image prompts (along with text descriptions).
The results show that the fine-tuned model generates significantly more varied static scenes compared to the source model, especially for cases with humans and animals.
The visualizations in \autoref{fig:supp1} illustrate that LoRA enables the generation of more varied static scenes without compromising visual quality.

To assess the role of LoRA in \textit{enhancing camera controllability}, we trained our full model without incorporating LoRA modules. 
For these scenarios, in the main branch, the camera embeddings are fed into the network in a channel-concatenation manner without LoRA tuning in the main backbone. 
Only the LoRA-camera encoder and linear processing layers at the top are learned.
Following the experimental setup outlined in \autoref{sec:abl_dualbranch} of the paper, we evaluated the models on 100 video clips sampled from RealEstate10K.
As shown in the middle part of \autoref{tab:camera_abl}, comparing the ``\emph{Dual w/o LoraModule}" and ``\emph{Dual-branch}" configurations reveals that LoRA plays a critical role in fine-tuning the main branch.
Excluding LoRA results in a noticeable performance drop in both visual quality and camera-guidance precision.

\begin{table}[b]
\centering
\vspace{-0.4cm}
    \caption{\textbf{Analysis of architectural designs in camera-guided video generation model.}   We report the performance for visual quality (FID and FVD) and pose control precision ($R_{\text{err}}$ and $T_{\text{err}}$) from models trained on the RealEstate10K dataset. The first part of the table is adopted from \autoref{tab:camera_table} in the paper.}
    \label{tab:camera_abl}    
   \resizebox{\columnwidth}{!}{%
    \begin{tabular}{lcccc}
    \toprule   
        Architecture & \multicolumn{4}{c}{Metrics} \\
         \cmidrule(lr){2-5}
         & FID $\downarrow$ & FVD $\downarrow$ & $R_{\text{err}}$ $\downarrow$ & $T_{\text{err}}$ $\downarrow$ \\
        \midrule
        Lora-branch & 19.02 & 212.74 & 0.102 & 0.157 \\
        Ctrl-branch & 18.75 & 205.45 & 0.058 & 0.104 \\
        Dual-branch & \textbf{17.22} & \textbf{183.54} & \textbf{0.052} & \textbf{0.095} \\
        \midrule
        Dual w/o LoraModule & 17.84 & 195.07 & 0.062 & 0.101 \\
        \midrule
        \textit{Ctrl-branch} only & \multicolumn{4}{c}{}  \\
        w/o weight copy & 18.92 & 206.75 & 0.065 & 0.108  \\
        block-1 & 19.90 & 214.66 & 0.114 & 0.162 \\
        blocks-10 & 19.15 & 210.74 & 0.075 & 0.126 \\
        blocks-30 & 20.15 & 221.61 & 0.056 & 0.105 \\
        \bottomrule
    \end{tabular}
    }
\end{table}

\subsection{Analysis of the ControlNet-Branch Design}

In our full model, the ControlNet branch is designed by utilizing a trainable copy of the first 21 base transformer blocks of the foundational video model, which consists of 42 blocks in total.
We extensively evaluated the design by using the ControlNet branch only.
Specifically, we trained the model with ControlNet conditioning under various configurations, including using the first 21 blocks without weight copying, denoted \emph{w/o weight copy}, and using the first 1, 10, or 30 blocks with weight copying, denoted block-1, blocks-10, and blocks-30, respectively.
As shown in the third part of \autoref{tab:camera_abl}, comparisons among the different architectural variants and the``\emph{Ctrl-branch}" reveal that weight copying substantially improves all metrics, particularly visual quality.
Using only one block results in weak camera controllability, whereas increasing the number of blocks strengthens the ability of the model to guide camera poses.
Notably, using 21 blocks (``\emph{Ctrl-branch}") achieves similar levels of pose controllability as using 30 blocks, while maintaining high visual quality.
Based on these observations, we selected the trainable copy of the first 21 base transformer blocks, as it provides an optimal balance between pose controllability and computational efficiency.

\section{Additional Analysis of 3D Reconstruction}

This section provides further analysis underlying our design choices related to the large-scale 3D reconstruction model.

\subsection{Fine-Tuning With the In-the-Wild Dataset}

In deploying the LaLRM, we adopt a progressive training strategy. During the second stage, we fine-tune the model by involving a self-generated in-the-wild dataset. 
To assess the impact of this dataset, we further fine-tuned a separate reconstruction model, LaLRM--, which excludes the in-the-wild dataset.
We quantitatively compared the performances of LaLRM-- and LaLRM on benchmark datasets and qualitatively evaluated them on 20 disjoint in-the-wild image prompts. 
The results reported in \autoref{tab:3d_abl} indicate that incorporating the in-the-wild dataset during fine-tuning enhances the generalization capabilities of our model.
Furthermore, as shown in \autoref{fig:supp2}, LaLRM demonstrates noticeably better rendering quality compared to LaLRM--. This further validates the benefits of using in-the-wild data in the fine-tuning process.

\begin{table}
\centering
    \caption{\textbf{Analysis on involving in-the-wild dataset to fine-tune LaLRM.} We report the performance on various benchmark datasets for novel view synthesis of 3D scenes, which are built from \emph{single view} condition.}
    \label{tab:3d_abl}
    \resizebox{\columnwidth}{!}{
    \begin{tabular}{lccccccccc}
    \toprule     
 Method & \multicolumn{3}{c}{RealEstate10K} & \multicolumn{3}{c}{DL3DV} & \multicolumn{3}{c}{Tanks-and-Temples} \\
         \cmidrule(lr){2-4} \cmidrule(lr){5-7} \cmidrule(lr){8-10}
        \textit{Metrics} & LPIPS $\downarrow$ & PSNR $\uparrow$ & SSIM $\uparrow$ & LPIPS $\downarrow$ & PSNR $\uparrow$ & SSIM $\uparrow$ & LPIPS $\downarrow$ & PSNR $\uparrow$ & SSIM $\uparrow$ \\
        \midrule
        LaLRM-- & 0.295 & 17.06 & 0.538 & 0.343 & 16.62 & 0.570 & 0.359 & 15.85 & 0.502 \\
        \rowcolor[rgb]{ .816,  .808,  .808} LaLRM & \textbf{0.292} & \textbf{17.15} & \textbf{0.550} & \textbf{0.325} & \textbf{16.64} & \textbf{0.574} & \textbf{0.344} & \textbf{15.90} & \textbf{0.510} \\      
        \bottomrule
    \end{tabular}}
    % \vspace{-0.3cm}
\end{table}

\section{Additional Discussion about Related Work}

This section provides a detailed discussion of our work relative to prior research in the use of generative priors for 3D rendering.

Significant advancements have been achieved in static and dynamic 3D object generation from text or single image prompts with notable improvements in quality and efficiency~\citep{wang2023prolificdreamer, bahmani2023cc3d, xu2024agg, bahmani20244d, voleti2024sv3d, chai2024star, xie2024sv4d, shi2023mvdream, li2024vivid, xu2024comp4d, liang2024diffusion4d, chen2024v3d, yang2024diffusion, gu2023nerfdiff, han2024vfusion3d, he2024gvgen, jiang2024animate3d, long2024wonder3d, ren2024l4gm}.\footnote{The references cited in this document are listed in the main paper.}
However, progress in scene-level generation has lagged behind~\citep{cat3d,viewcrafter,realdreamer,zhao2024genxd,yu2024wonderjourney, yu2024wonderworld, szymanowicz2025bolt3d, lu2024genex,wang2024vistadream,chen2024liftimage3d}.
Most approaches to 3D scene generation follow a two-stage process: First, novel views are generated from a single image and, second, these views are used to train a 3D representation with a per-scene optimization strategy.

Early methods, such as LucidDreamer~\citep{lucid} and RealmDreamer~\citep{realdreamer}, explored scene-level 3D generation conditioned on text descriptions or single images. 
They relied on the 3D priors from incomplete point clouds constructed via depth prediction from single images. 
Then, they combined depth-based warping with diffusion-based image inpainting to complete the scenes in an autoregressive manner. 
These methods often struggle with inconsistencies in occluded regions, as the per-view inpainting process can introduce severe artifacts and discontinuities, particularly in unseen areas.
WonderJourney~\citep{yu2024wonderjourney}, which targets wide-scene generation, also employs image inpainting diffusion models to fill unseen regions rendered from limited point clouds. 
However, as shown in our main comparisons, this method similarly suffers from 3D incoherence in occluded areas. Also, all these works do not have automatic and explicit control over camera poses during the generation process.
Other efforts, such as Cat3D~\citep{cat3d} and ReconFusion~\citep{reconfusion}, address multi-view consistency by incorporating camera conditioning into image diffusion models. Nonetheless, a noticeable issue is the tendency to produce blurry or distorted background regions, particularly when conditioned on a single image. This arises from the use of image diffusion models to obtain dense views auto-regressively, which are then used for 3D reconstruction via per-scene optimization ~\citep{kim2023neuralfield}. Image diffusion models lack built-in mechanisms to guarantee cross-view consistency and such a multiple-shot generation strategy often introduces inconsistencies, especially for wide-view scenarios.

More recent efforts, such as ReconX~\citep{reconx} and ViewCrafter~\citep{viewcrafter}, leverage video diffusion models and global point clouds to enhance multi-view consistency. However, as demonstrated in our main comparisons, these methods are sensitive to the initialization of point clouds and are restricted to generating narrow-scope scene representations. Additionally, they lack explicit pose control during the generation process.

Importantly, all previous methods depend on time-consuming per-scene optimization such as NeRF~\citep{nerf} or 3DGS~\citep{3dgs}.
By contrast, our approach integrates explicit camera control into a video diffusion model to enable precise and expansive scene generation. Our large-scale 3D reconstruction model is capable of efficiently constructing 3D scenes from video latents.
Its design effectively aligns the generation and reconstruction tasks and bridges the image space and 3D space through the video latent space, eliminating the need for time-consuming per-scene optimization.

\section{Implementation Details}
\label{supp:implementation-details}

\autoref{Tab:notation} provides a list of the mathematical notation used in the paper.

\subsection{Additional Architectural Details}
\label{supp:model-architecture}

Our framework comprises a Camera-Guided Video Diffusion Model and a Latent Large Reconstruction Model (LaLRM) capable of generating 3D scenes conditioned on single images.
Our architectural designs effectively align the generation and reconstruction tasks, bridging the image space and 3D space through the video latent space.

\begin{figure*}
  \centering
  \includegraphics[width=\linewidth]{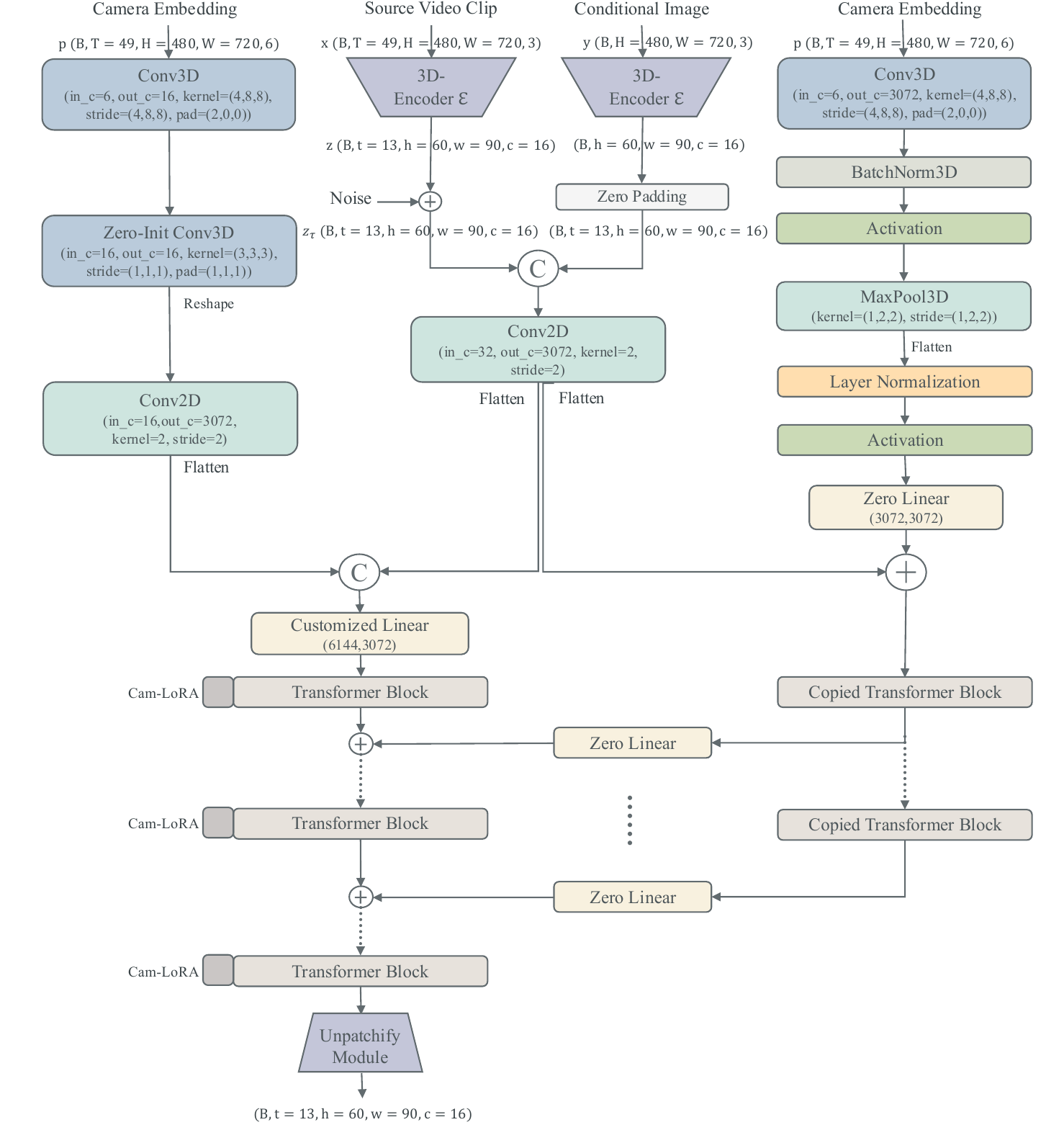}
  \caption{\textbf{Architecture of Dual-branch Camera-guided Video Diffusion Model.} We show the skeletons of the training pipeline, where random noise is added to the video latents. The conditional image is merged to the noisy latents via feature concatenation. The camera guidance is integrated with LoRA-branch (left) and ControlNet-branch (right). We ignore the text tokens, the diffusion time embeddings, the positional embeddings, and some reshaping operations for simplicity in the figure. In the foundation diffusion transformer, the text tokens are concatenated along \textit{number-of-token} dimension with visual tokens. Thus we apply zero-padding to camera tokens to guarantee the same length before concatenation or element-wise sum. By default, we use SiLu as our activation function.}
  \label{fig:video_struct}
\end{figure*}

\autoref{fig:video_struct} illustrates the architectural details of integrating camera embeddings into the pre-trained video diffusion transformer during the training stage, with a ControlNet-branch and a LoRA-branch.
Each branch involves a lightweight camera encoder composed of convolutional layers and zero-linear layers.
The camera encoders project the camera embedding into camera tokens of the same dimension as the visual tokens.
The visual tokens and camera tokens are concatenated or element-wise added before they are fed into the main branch and the ControlNet branch.
The visual-camera tokens are further processed by a sequence of transformer blocks and mapped to the same dimension of the added noise by the unpatchify module.
Note that for simplicity in \autoref{fig:video_struct}, we have omitted the text tokens, diffusion time embeddings, and positional embeddings.

\begin{figure}[h!]
  \centering
  \includegraphics[width=\linewidth]{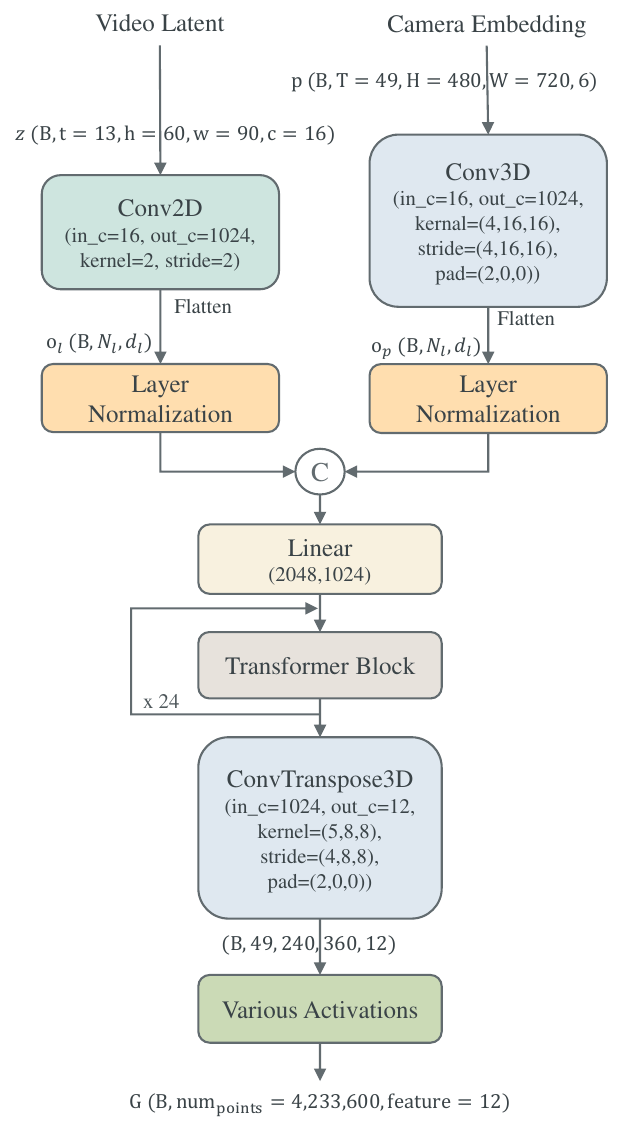}
  \caption{\textbf{Architecture of the Latent Large Reconstruction Model (LaLRM).} Given a video latent $z$ and the the Plücker embedding of cameras, the LaLRM directly regresses the 3DGS of the scenes in a feed-forward manner.}
  \label{fig:lalrm_struct}
\end{figure}

\autoref{fig:lalrm_struct} illustrates the architectural details of the LaLRM.
Given video latents that can be generated by the video diffusion model (during inference time) or can be embedded from source video clips (during training time), a lightweight tokenization module projects these latents to visual tokens.
Patchification is performed on the video latent along the spatial dimensions with a mild patch size $p_l$ to obtain $o_l\in R^{N_l\times d_l}$, where $N_l=t\cdot \frac{h}{p_l}\cdot \frac{w}{p_l}$.
For the Plücker embedding, we 3D-patchify along spatiotemporal dimensions with 3DConv. We purposefully set the spatial patch size to $p_l\cdot r_s$ and the temporal patch size to $r_t$ (defined in \autoref{sec:pre_video} of the paper), producing pose tokens $o_p$ that match the length with $o_l$.
The two sets of tokens are merged with channel-wise concatenation and fed into a sequence of transformer blocks to regress the 3D Gaussian features.
We fulfill pixel-level correspondence between the Gaussians and source video $x$ in the RGB space via the latent decoding module, which involves a 3D-DeConv layer with upsampling strides $(r_t, p_l\cdot r_s, p_l\cdot r_s)$ and the 12-channel output is the Gaussian feature map $G\in R^{(T\cdot H\cdot W)\times 12}$.

\subsection{Training and Evaluation Details}
\label{supp:training-detail}

The camera-guided video diffusion model uses a trans\-former-based video diffusion model; namely, CogVideoX-5B-I2V~\citep{cogvideo} that generates 49 frames of $480\times720$ pixel resolution.
An encoder from 3DVAE compresses video clips with ratios of $r_t=4$ and $r_s=8$, yielding latents of dimensionality $13\times 60\times 90$.
To build the ControlNet branch, we use the first $N=21$ base transformer blocks from the video backbone to initialize the weights.
The camera-LoRA has a low rank of dimension 256.
The model is trained with a batch size of 24 for 40K steps, using the Adam optimizer~\citep{diederik2014adam} with a learning rate of $2\times 10^{-5}$, $\beta_1=0.9$, $\beta_2=0.95$, and weight decay $1\times 10^{-4}$.

During the evaluation, we randomly sampled 300, 300, and 100 video clips from the RE10K test set, DL3DV-140, and Tanks-and-Temples, respectively.
For each video clip, we sampled a starting frame as the image condition and the subsequent $n$ camera poses as pose conditions, where $n$ is determined by the length of the generated video.
Since different datasets have diverse FPS, we sampled $n$ poses at different strides $s$ ($s=3$ for RE10K, $s=1$ for DL3DV-140, and $s=4$ for Tanks-and-Temples) to ensure smooth and noticeable view transitions.
Compared with the baselines, our method generates the longest videos, so the baselines perform inference using the front subset of our sampled poses.
The generated videos are COLMAPed to obtain camera poses, followed by conversion of the camera system to be relative to the first frame and normalization of all cameras to a common scale~\citep{he2024cameractrl, bahmani2024vd3d}.
We use translation and rotation error to measure camera-guidance precision~\citep{he2024cameractrl,bahmani2024vd3d,bahmani2024ac3d,xu2024camco}. For a fair comparison, we measure the mean errors across the first 16 frames for each method; \ie, VD3D~\citep{bahmani2024vd3d} (16-frame), ViewCrafter~\citep{viewcrafter} (25-frame), and Ours (49-frame).
SVD-based~\citep{blattmann2023stable} MotionCtrl~\citep{wang2024motionctrl} generates 14-frame videos, and all frames are used.
Visual similarity is assessed by calculating PSNR, SSIM~\citep{ssim}, and LPIPS~\citep{zhang2018unreasonable} between the generated images and ground-truth views.
Similarly, for a fair comparison, the first 14 frames in the generated videos are evaluated. An important issue is that generated videos tend to deviate from the conditional view and present diverse appearances as the scene progresses. It is therefore less reliable to evaluate quality using similarity metrics that measure the differences between generated frames and ground-truth views.

For the latent large reconstruction model, we use a patch size $p_l=2$ for the visual latent, and use a temporal patch size of 4 and a spatial patch size of 16 for the camera Plücker embedding.
We follow reference~\citep{gslrm} and use the same architecture for the transformer blocks. 
We use 24 base transformer blocks with a hidden dimension of 1,024.
The latent decoding module has a 3D-DeConv layer with upsampling strides (4, 16, 16).
The backbone transformer network is efficiently implemented with FlashAttentionV2~\citep{dao2023flashattention} and optimized with mixed precision training~\citep{micikevicius2017mixed} using the BF16 datatype.

We first train the model with low-resolution video clips of dimensionality $49\times 240\times 360$ and their corresponding latents of dimensionality $13\times 30\times 45$. 
Then, we fine-tune the model with high-resolution $49\times 480\times 720$ dimensional video clips and corresponding $13\times 60\times 90$ dimensional latents. 
At this stage, due to memory constraints, we modify the 3D-DeConv layer in the latent decoding module with upsampling strides (4, 8, 8).
Even with a smaller upsampling rate, for each 3D scene, our Gaussian prediction yields an enormous quantity of Gaussians to construct each scene; \ie, $T\times \frac{H}{2} \times \frac{W}{2}$ (4,233,600).
A total of $V=48$ supervision views are used, for which we randomly select $V'=24$ frames from each sampled video clip as seen views, and an additional 24 frames disjoint from the video clip as unseen views.
The model is trained on low-resolution and high-resolution datasets for 200K and 100K iterations, respectively, using a cosine annealing schedule at a peak learning rate of $4\times 10^{-4}$ and $1\times 10^{-5}$.
We use a batch size of 24 with the Adam optimizer with $\beta_1=0.9$, $\beta_2=0.95$, and weight decay $1\times 10^{-4}$. To improve the out-of-domain generalization capability, we generate 20K videos with image prompts from Flux.1~\citep{flux_repo} with camera poses sampled from RE10K.
These videos are used for training at the high-resolution setting.
This progressive training strategy is simple but effective in ensuring model convergence and improving out-of-domain generalizability.

In the evaluation stage, following the video generation, we sample starting frames and poses with different strides across multiple benchmark datasets for video generation.
The generated video latents are leveraged to reconstruct 3D scenes.
Since camera poses are required for reconstruction, we consider using the conditional camera poses directly and obtain descent results. 
Meanwhile, we empirically find that using COLMAP-derived poses from the generated videos yields superior results.
The rendered images from the constructed 3DGS are used for evaluation. 
At this stage, we evaluate performance exclusively using similarity-based metrics. Consistent with the video generation evaluation, we measure these metrics on the frames rendered from the first 14 camera poses.

\section{Limitations and Future Work}

Although our method achieves superior generation performance and higher efficiency relative to previous work, there remain some limitations.

First, the development of the model is time- and compute-intensive.
% Training both the camera-controlled video generation model and the 3D reconstruction model requires over 100 GPU-days.
Although the inference process is efficient, the inference speed of the video generation model remains a bottleneck. Most of the compute time in our pipeline is consumed during the video generation phase. This drawback could be mitigated via parallel computation; \eg, using xDiT ({\footnotesize\url{https://github.com/xdit-project/xDiT}}) for parallel inference or using a more efficient denoising strategy. 

Second, our approach focuses on static scenes. We occasionally observe motions in the generated videos, which hampers the reconstruction effect. In future work, we will aim to extend our pipeline to dynamic scenes, exploring its potential for generating 4D content incorporating temporal dynamics.

By addressing the limitations mentioned above, our framework can be improved for broader applications with enhanced performance.

\end{document}